# Binary Encodings of Non-binary Constraint Satisfaction Problems: Algorithms and Experimental Results

**Nikolaos Samaras**                                            SAMARAS@UOM.GR
*Department of Applied Informatics*
*University of Macedonia, Greece*

**Kostas Stergiou**                                             KONSTERG@AEGEAN.GR
*Department of Information and Communication Systems Engineering*
*University of the Aegean, Greece*

## Abstract

A non-binary Constraint Satisfaction Problem (CSP) can be solved directly using extended versions of binary techniques. Alternatively, the non-binary problem can be translated into an equivalent binary one. In this case, it is generally accepted that the translated problem can be solved by applying well-established techniques for binary CSPs. In this paper we evaluate the applicability of the latter approach. We demonstrate that the use of standard techniques for binary CSPs in the encodings of non-binary problems is problematic and results in models that are very rarely competitive with the non-binary representation. To overcome this, we propose specialized arc consistency and search algorithms for binary encodings, and we evaluate them theoretically and empirically. We consider three binary representations; the hidden variable encoding, the dual encoding, and the double encoding. Theoretical and empirical results show that, for certain classes of non-binary constraints, binary encodings are a competitive option, and in many cases, a better one than the non-binary representation.

## 1. Introduction

Constraint Satisfaction Problems (CSPs) appear in many real-life applications such as scheduling, resource allocation, timetabling, vehicle routing, frequency allocation, etc. Most CSPs can be naturally and efficiently modelled using non-binary (or n-ary) constraints that may involve an arbitrary number of variables. It is well known that any non-binary CSP can be converted into an equivalent binary one. The most well-known translations are the dual encoding (Dechter & Pearl, 1989) and the hidden variable encoding (Rossi, Petrie, & Dhar, 1990). The ability to translate any non-binary CSP into binary has been often used in the past as a justification for restricting attention to binary CSPs. Implicitly, the assumption had been that when faced with a non-binary CSP we can simply convert it into a binary one, and then apply well-known generic techniques for solving the binary equivalent. In this paper we will show that this assumption is flawed because generic techniques for binary CSPs are not suitable for binary encodings of non-binary problems.

In the past few years, there have been theoretical and empirical studies on the efficiency of binary encodings and comparisons between binary encodings and the non-binary representation (Bacchus & van Beek, 1998; Stergiou & Walsh, 1999; Mamoulis & Stergiou, 2001; Smith, 2002; Bacchus, Chen, van Beek, & Walsh, 2002). Theoretical results have showed that converting non-binary CSPs into binary equivalents is a potentially efficient way to solve certain classes of non-binary problems. However, in the (limited) empirical studies there are very few cases where this appears to be true, with Conway's "game of Life" (Smith, 2002) being the most notable exception. There are various reasons for this. In many cases, the extensive space requirements of the binary encodings make them infeasible. Also, in many non-binary problems we can utilize efficient specialized propagators for certain constraints, such as the algorithm developed by Régin (1994) for the all-different constraint. Converting such constraints into binary is clearly impractical. Another reason, which has been overlooked, is that most (if not all) experimental studies use well-known generic local consistency and search algorithms in the encodings. In this way they fail to exploit the structure of the constraints in the encodings, ending up with inefficient algorithms. To make the binary encodings a realistic choice of modelling and solving non-binary CSPs, we need algorithms that can utilize their structural properties. Finally, it is important to point out that the use of a binary encoding does not necessarily mean that we have to convert all the non-binary constraints in a problem into binary, as it is commonly perceived. If we are selective in the constraints we encode, based on properties such as arity and tightness, then we can get efficient hybrid models.

To address these issues, we show that the use of specialized arc consistency and search algorithms for binary encodings of non-binary CSPs can lead to efficient models. We consider three encodings; the dual, the hidden variable, and the double encoding. The latter, which is basically the conjunction of the other two encodings, has received little attention but may well turn out to be the most significant in practice. The aim of this study is twofold. First, to present efficient algorithms for the binary encodings and analyze them theoretically and experimentally. Second, and more importantly, to investigate if and when the use of such algorithms can help solve non-binary problems efficiently. Towards these aims, we make the following contributions:

- We describe a simple algorithm that enforces arc consistency on the hidden variable encoding of an arbitrary non-binary CSP with $O(ekd^k)$ time complexity, where $e$ is the number of constraints, $k$ the maximum arity of the constraints, and $d$ the maximum domain size. This gives an $O(d)$ improvement compared to the asymptotic complexity of a generic arc consistency algorithm. The improved complexity is now the same as the complexity of an optimal generalized arc consistency algorithm in the non-binary representation of the problem. We also identify a property of the arc consistency algorithm for the hidden variable encoding that can make it run faster, on arc inconsistent problems, than the generalized arc consistency algorithm.

- We then consider search algorithms that maintain local consistencies during search in the hidden variable encoding. We show that, like maintaining arc consistency, all the generalizations of forward checking to non-binary CSPs can be emulated by corresponding forward checking algorithms that run in the hidden variable encoding and only instantiate original variables (i.e. the variables of the initial non-binary problem). We show that each such algorithm and its corresponding algorithm for non-binary constraints have the following relationships: 1) they visit the same number of search tree nodes, and 2) the asymptotic cost of each of them is within a polynomial bound of the other.

- We describe a specialized algorithm for the dual encoding that achieves arc consistency with $O(e^3 d^k)$ worst-case time complexity. This is significantly lower than the $O(e^2 d^{2k})$ complexity of a generic arc consistency algorithm. The improvement in the complexity bound stems from the observation that constraints in the dual encoding have a specific structure; namely they are piecewise functional (Van Hentenryck, Deville, & Teng, 1992). Apart from applying arc consistency in the dual encoding of a non-binary CSP, this algorithm can also be used as a specialized filtering algorithm for certain classes of non-binary constraints.

- We adapt various search algorithms to run in the double encoding and compare them theoretically to similar algorithms for the hidden variable encoding and the non-binary representation. Search algorithms that operate in the double encoding can exploit the advantages of both the hidden variable and the dual encoding. For example, we show that, under certain conditions, the asymptotic cost of the maintaining arc consistency algorithm in the double encoding can only be polynomially worse than the asymptotic cost of the corresponding algorithm in the non-binary representation (and the hidden variable encoding), while it can be exponentially better.

- Finally, we make an extensive empirical study on various domains. We consider random problems as well as structured ones, like crossword puzzle generation, configuration, and frequency assignment. This study consists of two parts. In the first part, we give experimental results that demonstrate the advantages of the specialized algorithms for binary encodings compared to generic algorithms. For example, the specialized arc consistency algorithm for the dual encoding can be orders of magnitude faster than a generic arc consistency algorithm. In the second part we show that the use of binary encodings can offer significant benefits when solving certain classes of non-binary CSPs. For example, solving the dual encoding of some configuration problems can be orders of magnitudes more efficient than solving the non-binary representation. Also, empirical results from frequency assignment - like problems demonstrate that a binary encoding can be beneficial even for non-binary constraints that are intentionally specified.

This paper is structured as follows. In Section 2 we give the necessary definitions and background. In Section 3 we describe a specialized arc consistency algorithm for the hidden variable encoding. We also demonstrate that all the extensions of forward checking to non-binary CSPs can be emulated by binary forward checking algorithms that run in the hidden variable encoding. In Section 4 we explain how the complexity of arc consistency in the dual encoding can be improved and describe a specialized arc consistency algorithm. Section 5 discusses algorithms for the double encoding. In Section 6 we present experimental results on random and structured problems that demonstrate the usefulness of the proposed algorithms. We also draw some conclusions regarding the applicability of the encodings, based on theoretical and experimental results. Section 7 discusses related work. Finally, in Section 8 we conclude.

## 2. Background

In this section we give some necessary definitions on CSPs, and describe the hidden variable, dual, and double encodings of non-binary CSPs.

### 2.1 Basic Definitions

A *Constraint Satisfaction Problem* (CSP), $P$, is defined as a tuple $(X, D, C)$, where:

- $X = \{x_1, \ldots, x_n\}$ is a finite set of $n$ variables.

- $D = \{D_{in}(x_1), \ldots, D_{in}(x_n)\}$ is a set of initial domains. For each variable $x_i \in X$, $D_{in}(x_i)$ is the initial finite domain of its possible values. CSP algorithms remove values from the domains of variables through value assignments and propagation. For any variable $x_i$, we denote by $D(x_i)$ the current domain of $x_i$ that at any time consists of values that have not been removed from $D_{in}(x_i)$. We assume that for every $x_i \in X$, a total ordering $<_d$ can be defined on $D_{in}(x_i)$.

- $C = \{c_1, \ldots, c_e\}$ is a set of $e$ constraints. Each constraint $c_i \in C$ is defined as a pair $(vars(c_i), rel(c_i))$, where 1) $vars(c_i) = \{x_{j_1}, \ldots, x_{j_k}\}$ is an ordered subset of $X$ called the constraint *scheme*, 2) $rel(c_i)$ is a subset of the *Cartesian* product $D_{in}(x_{j_1}) \times \ldots \times D_{in}(x_{j_k})$ that specifies the allowed combinations of values for the variables in $vars(c_i)$.

The size of $vars(c_i)$ is called the *arity* of the constraint $c_i$. Constraints of arity 2 are called *binary*. Constraints of arity greater than 2 are called *non-binary* (or *n-ary*). Each tuple $\tau \in rel(c_i)$ is an ordered list of values $(a_1, \ldots, a_k)$ such that $a_j \in D_{in}(x_j), j = 1, \ldots, k$. A tuple $\tau = (a_1, \ldots, a_k)$ is *valid* if $\forall a_j, j \in 1, \ldots, k, a_j \in D(x_j)$. That is, a tuple is valid if all the values in the tuple are present in the domains of the corresponding variables. The process which verifies whether a given tuple is allowed by a constraint $c_i$ or not is called a *consistency check*. A constraint can be either defined *extensionally* by the set of allowed (or disallowed) tuples or *intensionally* by a predicate or arithmetic function. A binary CSP can be represented by a graph (called constraint graph) where nodes correspond to variables and edges correspond to constraints. A non-binary CSP can be represented by a constraint hyper-graph where the constraints correspond to hyper-edges connecting two or more nodes.

The assignment of value $a$ to variable $x_i$ will be denoted by $(x_i, a)$. Any tuple $\tau = (a_1, \ldots, a_k)$ can be viewed as a set of value to variable assignments $\{(x_1, a_1), \ldots, (x_k, a_k)\}$. The set of variables over which a tuple $\tau$ is defined will be denoted by $vars(\tau)$. For any subset $vars'$ of $vars(\tau)$, $\tau[vars']$ denotes the sub-tuple of $\tau$ that includes only assignments to the variables in $vars'$. Any two tuples $\tau$ and $\tau'$ of $rel(c_i)$ can be ordered by the lexicographic ordering $<_{lex}$. In this ordering, $\tau <_{lex} \tau'$ iff there a exists a subset $\{x_1, \ldots, x_j\}$ of $c_i$ such that $\tau[x_1, \ldots, x_j] = \tau'[x_1, \ldots, x_j]$ and $\tau[x_{j+1}] <_{lex} \tau'[x_{j+1}]$. An assignment $\tau$ is *consistent*, if for all constraints $c_i$, where $vars(c_i) \subseteq vars(\tau)$, $\tau[vars(c_i)] \in rel(c_i)$. A *solution* to a CSP $(X, D, C)$ is a consistent assignment to all variables in $X$. If there exists a solution for a given CSP, we say that the CSP is *soluble*. Otherwise, it is *insoluble*.

A basic way of solving CSPs is by using backtracking search. This can be seen as a traversal of a *search tree* which comprises of the possible assignments of values to variables.

Each level of the tree corresponds to a variable. A node in the search tree corresponds to a tuple (i.e. an assignment of values to variables). The root of the tree corresponds to the empty tuple, the first level nodes correspond to 1-tuples (an assignment of a value to one variable), the second level nodes correspond to 2-tuples (assignment of values to two variables generated by extending the first level 1-tuples) etc. At each stage of the search tree traversal, the variables that have been already assigned are called *past variables*. The most recently assigned variable is called *current variable*. The variables that have not been assigned yet are called *future variables*.

In the rest of this paper we will use the notation $n$ for the number of variables in a CSP, $e$ for the number of constraints in the problem, $d$ for the maximum domain size of the variables, and $k$ for the maximum arity of the constraints.

### 2.1.1 Arc Consistency

An important concept in CSPs is the concept of local consistency. Local consistencies are properties that can be applied in a CSP, using (typically) polynomial algorithms, to remove inconsistent values either prior to or during search. Arc consistency is the most commonly used local consistency property in the existing constraint programming engines. We now give a definition of arc consistency.

**Definition 2.1** A value $a \in D(x_j)$ is consistent with a constraint $c_i$, where $x_j \in vars(c_i)$ if $\exists \tau \in rel(c_i)$ such that $\tau[x_j] = a$ and $\tau$ is valid. In this case we say that $\tau$ is a support of $a$ in $c_i$. A constraint $c_i$ is *Arc Consistent* (AC) iff for each variable $x_j \in vars(c_i)$, $\forall\, a \in D(x_j)$, there exists a support for $a$ in $c_i$. A CSP $(X, D, C)$ is arc consistent iff there is no empty domain in $D$ and all the constraints in $C$ are arc consistent.

Arc consistency can be enforced on a CSP by removing all the unsupported values from the domains of variables. By *enforcing* arc consistency (or some local consistency property $A$ in general) on a CSP $P$, we mean applying an algorithm that yields a new CSP that is arc consistent (or has the property $A$) and has the same set of solutions as $P$. The above definition of arc consistency applies to constraints of any arity. To distinguish between the binary and non-binary cases, we will use the term arc consistency (AC) to refer to the property of arc consistency for binary constraints only. For non-binary constraints we will use the term *Generalized Arc Consistency* (GAC).

The usefulness of AC processing was recognized early, and as a result, various AC algorithms for binary constraints have been proposed in the literature (e.g. AC-3 in Mackworth, 1977, AC-4 in Mohr & Henderson, 1986, AC-5 in Van Hentenryck et al., 1992, AC-7 in Bessière et al., 1995, AC-2001 in Bessière & Régin, 2001, AC3.1 in Zhang & Yap, 2001). Some of them have been extended to the non-binary case (e.g. GAC-4 in Mohr & Masini, 1988, GAC-Schema in Bessière & Régin, 1996a, GAC-2001 in Bessière & Régin, 2001). AC can be enforced on a binary CSP with $O(ed^2)$ optimal worst-case time complexity. The worst-case complexity of enforcing GAC on a non-binary CSP is $O(ekd^k)$ (Bessière & Régin, 1996a).

In this paper we use algorithms AC-2001 and GAC-2001 for theoretical and empirical comparisons with specialized algorithms for the encodings. This is not restrictive, in the sense that any generic AC (and GAC) algorithm can be used instead.

Following Debruyne & Bessiére (2001), we call a local consistency property $A$ *stronger* than $B$ iff for any problem enforcing $A$ deletes at least the same values as $B$, and *strictly stronger* iff it is stronger and there is at least one problem where $A$ deletes more values than $B$. We call $A$ *equivalent* to $B$ iff they delete the same values for all problems. Similarly, we call a search algorithm $A$ stronger than a search algorithm $B$ iff for every problem $A$ visits at most the same search tree nodes as $B$, and strictly stronger iff it is stronger and there is at least one problem where $A$ visits less nodes than $B$. $A$ is equivalent to $B$ iff they visit the same nodes for all problems.

Following Bacchus et al. (2002), the *asymptotic cost* (or just cost hereafter) of a search algorithm $A$ is determined by the worst-case number of nodes that the algorithm has to visit to solve the CSP, and the worst-case time complexity of the algorithm at each node[1]. As in the paper by Bacchus et al. (2002), we use this measure to set asymptotic bounds in the relative performance of various algorithms. For example, if two algorithms $A$ and $B$ always visit the same nodes and $A$ enforces a property at each node with exponentially higher complexity than the property enforced by $B$, then we say that *algorithm $A$ can have an exponentially greater cost than algorithm $B$*.

### 2.1.2 Functional and Piecewise Functional Constraints

The specialized AC algorithms for the hidden variable and the dual encoding that we will describe in Sections 3 and 4 exploit structural properties of the encodings. As we will explain in detail later, the binary constraints in the hidden variable encoding are one-way functional, while the binary constraints in the dual encoding are piecewise functional. We now define these concepts.

**Definition 2.2** A binary constraint $c$, where $vars(c) = \{x_i, x_j\}$, is *functional* with respect to $D(x_i)$ and $D(x_j)$ iff for all $a \in D(x_i)$ (resp. $b \in D(x_j)$) there exists at most one value $b \in D(x_j)$ (resp. $a \in D(x_i)$) such that $b$ is a support of $a$ in $c$ (resp. $a$ is a support of $b$).

An example of a functional constraint is $x_i = x_j$. A binary constraint is *one-way functional* if the functionality property holds with respect to only one of the variables involved in the constraint.

Informally, a piecewise functional constraint over variables $x_i$, $x_j$ is a constraint where the domains of $x_i$ and $x_j$ can be partitioned into groups such that each group of $D(x_i)$ is supported by at most one group of $D(x_j)$, and vice versa. To give a formal definition, we first define the concept of a piecewise decomposition.

**Definition 2.3** (Van Hentenryck et al., 1992) Let $c$ be a binary constraint with $vars(c) = \{x_i, x_j\}$. The partitions $S = \{s_1, \ldots, s_m\}$ of $D(x_i)$ and $S' = \{s'_1, \ldots, s'_{m'}\}$ of $D(x_j)$ are a *piecewise decomposition* of $D(x_i)$ and $D(x_j)$ with respect to $c$ iff for all $s_l \in S, s'_{l'} \in S'$, the following property holds: either for all $a \in s_l, b \in s'_{l'}, (a, b) \in rel(c)$, or for all $a \in s_l, b \in s'_{l'}, (a, b) \notin rel(c)$.

---

1. In the paper by Bacchus et al. (2002) the cost of applying a variable ordering heuristic at each node is also taken into account. When we theoretically compare search algorithms in this paper we assume that they use the same variable ordering, so we do not take this cost into account.

**Definition 2.4** (Van Hentenryck et al., 1992) A binary constraint $c$, where $vars(c) = \{x_i, x_j\}$, is *piecewise functional* with respect to $D(x_i)$ and $D(x_j)$ iff there exists a piecewise decomposition $S = \{s_1, \ldots, s_m\}$ of $D(x_i)$ and $S' = \{s'_1, \ldots, s'_{m'}\}$ of $D(x_j)$ with respect to $c$ such that for all $s_l \in S$ (resp. $s'_{l'} \in S'$), there exists at most one $s'_{l'} \in S'$ (resp. $s_l \in S$), such that for all $a \in s_l$, $b \in s'_{l'}$ $(a,b) \in rel(c)$.

Example of piecewise functional constraints are the modulo ($x_2$ MOD $x_3 = a$) and integer division ($x_2$ DIV $x_3 = a$) constraints.

### 2.2 Binary Encodings

There are two well-known methods for transforming a non-binary CSP into a binary one; the dual graph encoding and the hidden variable encoding. Both encode the non-binary constraints to variables that have as domains the valid tuples of the constraints. That is, by building a binary encoding of a non-binary constraint we store the extensional representation of the constraint (the set of allowed tuples). A third method is the double encoding which combines the other two.

#### 2.2.1 Dual Encoding

The *dual encoding* (originally called *dual graph encoding*) was inspired by work in relational databases. In the dual encoding (DE) (Dechter & Pearl, 1989) the variables are swapped with constraints and vice versa. Each constraint $c$ of the original non-binary CSP is represented by a variable which we call a *dual variable* and denote by $v_c$. We refer to the variables of the original non-binary CSP as *original variables*. The domain of each dual variable $v_c$ consists of the set of allowed tuples in the original constraint $c$. Binary constraints between two dual variables $v_c$ and $v_{c'}$ exist iff $vars(c) \cap vars(c') \neq \emptyset$. That is, iff the constraints $c$ and $c'$ share one or more original variables. If *common_vars* is the set of original variables common to $c$ and $c'$ then a tuple $\tau \in D(v_c)$ is supported in the constraint between $v_c$ and $v_{c'}$ iff there exists a tuple $\tau' \in D(v_{c'})$ such that $\tau[common\_vars] = \tau'[common\_vars]$.

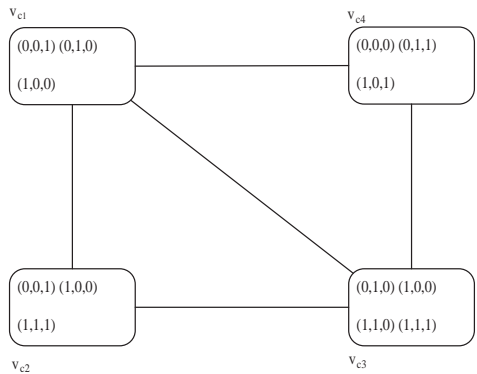

Figure 1: Dual encoding of a non-binary CSP.

Consider the following example with six variables with 0-1 domains, and four constraints: $c_1 : x_1 + x_2 + x_6 = 1$, $c_2 : x_1 - x_3 + x_4 = 1$, $c_3 : x_4 + x_5 - x_6 \geq 1$, and $c_4 : x_2 + x_5 - x_6 = 0$. The DE represents this problem with 4 dual variables, one for each constraint. The domains of these dual variables are the tuples that satisfy the respective constraint. For example, the dual variable $v_{c_3}$ associated with the third constraint has the domain $\{(0,1,0),(1,0,0),(1,1,0),(1,1,1)\}$ as these are the tuples of values for $(x_4, x_5, x_6)$ which satisfy $x_4 + x_5 - x_6 \geq 1$. As a second example, the dual variable $v_{c_4}$ associated with the last constraint has the domain $\{(0,0,0),(0,1,1),(1,0,1)\}$. Between $v_{c_3}$ and $v_{c_4}$ there is a compatibility constraint to ensure that the two original variables in common, $x_5$ and $x_6$, have the same values. This constraint allows only those pairs of tuples which agree on the second and third elements (i.e. $(1,0,0)$ for $v_{c_3}$ and $(0,0,0)$ for $v_{c_4}$, or $(1,1,1)$ for $v_{c_3}$ and $(0,1,1)$ for $v_{c_4}$). The DE of the problem is shown in Figure 1.

In the rest of this paper, we will sometimes denote by $c_{v_i}$ the non-binary constraint that is encoded by dual variable $v_i$. For an original variable $x_j \in vars(c_{v_i})$, $pos(x_j, c_{v_i})$ will denote the position of $x_j$ in $c_{v_i}$. For instance, given a constraint $c_{v_i}$ on variables $x_1, x_2, x_3$, $pos(x_2, c_{v_i}) = 2$.

### 2.2.2 Hidden Variable Encoding

The *hidden variable encoding* (HVE) was inspired by the work of philosopher Peirce (1933). According to Rossi et al. (1990), Peirce first showed that binary relations have the same expressive power as non-binary relations.

In the HVE (Rossi et al., 1990), the set of variables consists of all the original variables of the non-binary CSP plus the set of dual variables. As in the dual encoding, each dual variable $v_c$ corresponds to a constraint $c$ of the original problem. The domain of each dual variable consists of the tuples that satisfy the original constraint. For every dual variable $v_c$, there is a binary constraint between $v_c$ and each of the original variables $x_i$ such that $x_i \in vars(c)$. A tuple $\tau \in D(v_c)$ is supported in the constraint between $v_c$ and $x_i$ iff there exists a value $a \in D(x_i)$ such that $\tau[x_i] = a$.

Consider the previous example with six variables with 0-1 domains, and four constraints: $c_1 : x_1 + x_2 + x_6 = 1$, $c_2 : x_1 - x_3 + x_4 = 1$, $c_3 : x_4 + x_5 - x_6 \geq 1$, and $c_4 : x_2 + x_5 - x_6 = 0$. In the HVE there are, in addition to the original six variables, four dual variables. As in the DE, the domains of these variables are the tuples that satisfy the respective constraint. There are now compatibility constraints between a dual variable $v_c$ and the original variables contained in constraint $c$. For example, there are constraints between $v_{c_3}$ and $x_4$, between $v_{c_3}$ and $x_5$ and between $v_{c_3}$ and $x_6$, as these are the variables involved in constraint $c_3$. The compatibility constraint between $c_{v_3}$ and $x_4$ is the relation that is true iff the first element in the tuple assigned to $c_{v_3}$ equals the value of $x_4$. The HVE is shown in Figure 2.

### 2.2.3 Double Encoding

The *double encoding* (Stergiou & Walsh, 1999) combines the hidden variable and the dual encoding. As in the HVE, the set of variables in the double encoding consists of all the variables of the original non-binary CSP plus the dual variables. For every dual variable $v_c$, there is a binary constraint between $v_c$ and each of the original variables $x_i$ involved in the corresponding non-binary constraint $c$. As in the DE, there are also binary constraints between two dual variables $v_c$ and $v_{c'}$ if the non-binary constraints $c$ and $c'$ share one or more original variables.

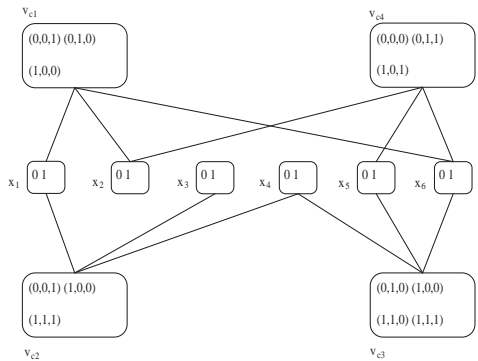

Figure 2: Hidden variable encoding of a non-binary CSP.

## 3. Algorithms for the Hidden Variable Encoding

In this section we discuss specialized algorithms for the HVE. We first describe a simple AC algorithm for the HVE that has the same worst-case time complexity as an optimal GAC algorithm for the non-binary representation. In Appendix A, we also show that for any arc consistent CSP the proposed AC algorithm performs exactly the same number of consistency checks as the corresponding GAC algorithm. For arc inconsistent problems we show that the AC algorithm for the HVE can detect the inconsistency earlier and thus perform fewer consistency checks than the GAC algorithm.

We also consider search algorithms for the HVE that maintain local consistencies during search. We show that, like maintaining arc consistency, the generalizations of forward checking to non-binary CSPs can be emulated by corresponding binary forward checking algorithms in the HVE that only instantiate original variables.

### 3.1 Arc Consistency

It has been proved that AC in the HVE is equivalent to GAC in the non-binary problem (Stergiou & Walsh, 1999). Since the HVE is a binary CSP, one obvious way to apply AC is by using a generic AC algorithm. However, this results in redundant processing and an asymptotic time complexity worse than $O(ekd^k)$. To be precise, in the HVE of a problem with $k-$ary constraints we have $ek$ binary constraints between dual and original variables. On such a constraint, AC can be enforced with $O(dd^k)$ worst-case time complexity. For the whole problem the complexity is $O(ekd^{k+1})$.

Instead, we will now describe a simple AC algorithm that operates in the HVE and achieves the same worst-case time complexity as an optimal GAC algorithm applied in the non-binary representation. We can achieve this by slightly modifying the GAC algorithm of Bessiére and Régin (2001) (GAC-2001). In Figure 3 we sketch the AC algorithm for the HVE, which we call HAC (Hidden AC).

**function** $HAC$
1:   $Q \leftarrow \emptyset$
2:   **for** each dual variable $v_j$
3:      **for** each variable $x_i$ where $x_i \in vars(c_{v_j})$
4:         **if** $Revise(x_i, v_j) = TRUE$
5:            **if** $D(x_i)$ is empty **return** INCONSISTENCY
6:            put in $Q$ each dual variable $v_l$ such that $x_i \in vars(c_{v_l})$
7:   **return** $Propagation$

**function** $Propagation$
8:   **while** $Q$ is not empty
9:      pop dual variable $v_j$ from $Q$
10:     **for** each unassigned variable $x_i$ where $x_i \in vars(c_{v_j})$
11:        **if** $Revise(x_i, v_j) = TRUE$
12:           **if** $D(x_i)$ is empty **return** INCONSISTENCY
13:           put in $Q$ each dual variable $v_l$ such that $x_i \in vars(c_{v_l})$
14:  **return** CONSISTENCY

**function** $Revise(x_i, v_j)$
15:  DELETION $\leftarrow$ FALSE
16:     **for** each value $a \in D(x_i)$
17:        **if** $currentSupport_{x_i,a,v_j}$ is not $valid$
18:           **if** $\exists \tau (\in D(v_j)) >_{lex} currentSupport_{x_i,a,v_j}$, $\tau[x_i] = a$ and $\tau$ is $valid$
19:              $currentSupport_{x_i,a,v_j} \leftarrow \tau$
20:           **else**
21:              remove $a$ from $D(x_i)$
22:              **for** each $v_l$ such that $x_i \in vars(c_{v_l})$
23:                 remove from $D(v_l)$ each tuple $\tau'$ such that $\tau'[x_i] = a$
24:                 **if** $D(v_l)$ is empty **return** INCONSISTENCY
25:              DELETION $\leftarrow$ TRUE
26:  **return** DELETION

Figure 3: HAC: an AC algorithm for the hidden variable encoding.

The HAC algorithm uses a stack (or queue) of dual variables to propagate value deletions, and works as follows. In an initialization phase it iterates over each dual variable $v_j$ (line 2). For every original variable $x_i$ constrained with $v_j$ the algorithm *revises* the constraint between $v_j$ and $x_i$. This is done by calling function *Revise* (line 4). During each revision, for each value $a$ of $D(x_i)$ we look for a tuple in the domain of $v_j$ that supports it. As in AC-2001, we store $currentSupport_{x_i,a,v_j}$: the most recent tuple we have found in $D(v_j)$ that supports value $a$ of variable $x_i$[2]. If this tuple has not been deleted from $D(v_j)$ then we know that $a$ is supported. Otherwise, we look for a new supporting tuple starting from the tuple immediately after $currentSupport_{x_i,a,v_j}$. If no such tuple is found then $a$ is removed from $D(x_i)$ (line 21). In that case, all tuples that include that value are removed from the domains of the dual variables that are constrained with $x_i$ (lines 22–23). If these dual variables are not already in the stack they are added to it[3]. Then, dual variables are removed from the stack sequentially. For each dual variable $v_j$ that is removed from the stack, the algorithm revises the constraint between $v_j$ and each original variable $x_i$ constrained with $v_j$. The algorithm terminates if all the values in a domain are deleted, in which case the problem is not arc consistent, or if the stack becomes empty, in which case the problem is arc consistent.

---

2. We assume, without loss of generality, that the algorithm looks for supports by checking the tuples in lexicographic order.

The main difference between HAC and GAC-2001 is that GAC-2001 does not include lines 22–24. That is, even if the non-binary constraints are given in extension, GAC-2001 does not remove tuples that become invalid from the lists of allowed tuples. As a result, the two algorithms check for the validity of a tuple (in lines 17 and 18) in different ways. Later on in this section we will explain this in detail. Apart from this difference, which is important because it affects their run times, the two algorithms are essentially the same. We can move from HAC to GAC-2001 by removing lines 22–24 and substituting any references to dual variables by references to the corresponding constraints. For example, $currentSupport_{x_i,a,v_j}$ corresponds to $currentSupport_{x_i,a,c_{v_j}}$ in GAC-2001, i.e. the last tuple in constraint $c_{v_j}$ that supports value $a$ of variable $x_i$. Note that in such an implementation of GAC-2001, propagation is *constraint-based*. That is, the algorithm utilizes a stack of constraints to perform the propagation of value deletions.

### 3.1.1 Complexities

We now give a upper bound on the number of consistency checks performed by HAC in the worst-case. Function $Revise(x_i,v_j)$ can be called at most $kd$ times for each dual variable $v_j$, once for every deletion of a value from the domain of $x_i$, where $x_i$ is one of the $k$ original variables constrained with $v_j$. In each call to $Revise(x_i,v_j)$ the algorithm performs at most $d$ checks (one for each value $a \in D(x_i)$) to see if $currentSupport_{x_i,a,v_j}$ is valid (line 17). If $currentSupport_{x_i,a,v_j}$ is not valid, HAC tries to find a new supporting tuple for $a$ in $D(v_j)$. To check if a tuple $\tau$ that contains the assignment $(x_i,a)$ supports $a$ we need to check if $\tau$ is valid. If a tuple is not valid then one of its values has been removed from the domain of the corresponding variable. This means that the tuple has also been removed from the domain of the dual variable. Therefore, checking the validity of a tuple can be done in constant time by looking in the domain of the dual variable. The algorithm only needs to check for support the $d^{k-1}$, at maximum, tuples that contain the assignment $(x_i,a)$. Since HAC stores $currentSupport_{x_i,a,v_j}$, at each call of $Revise(x_i,v_j)$ and for each value $a \in D(x_i)$, it only checks tuples that have not been checked before. In other words, we can check each of the $d^{k-1}$ tuples at most once for each value of $x_i$. So overall, in the worst case, we have $d^{k-1}$ checks plus the $d$ checks to test the validity of the current support. For $kd$ values the upper bound in checks performed by HAC to make one dual variable AC is $O(kd(d+d^{k-1}))=O(kd^k)$. For $e$ dual variables the worst-case complexity bound is $O(ekd^k)$, which is the same as the complexity of GAC in the non-binary representation.

---

3. Note that dual variables that are already in the stack are never added to it. In this sense, the stack is implemented as a set.

The asymptotic space complexity of the HAC algorithm is dominated by the $O(ed^k)$ space needed to store the domains of the dual variables. The algorithm also requires $O(nde)$ space to store the current supports. Since the space required grows exponentially with the arity of the constraints, it is reasonable to assume that the HVE (and the other binary encodings) cannot be practical for constraints of large arity, unless the constraints are very tight.

As mentioned, a consistency check in the non-binary representation is done in a different way than in the HVE. Assume that GAC-2001 looks for a support for value $a_i \in D(x_i)$ in constraint $c$, where $vars(c) = \{x_1, \ldots, x_k\}$ and $x_i \in vars(c)$. A tuple $\tau = (a_1, \ldots, a_k)$ supports $a_i$ if $\tau[x_i] = a_i$ and $\tau$ is valid. To check if $\tau$ is valid, GAC-2001 has to check if values $a_1, \ldots, a_k$ (except $a_i$) are still in the domains of variables $x_1, \ldots, x_k$. Therefore, in the worst case, a consistency check by GAC-2001 involves $k - 1$ operations. In contrast, HAC checks for the validity of a tuple in constant time by looking in the domain of the corresponding dual variable to see if the tuple is still there. However, this means that the algorithm has to update the (usually) large domains of the dual variables after a value deletion from an original variable. This affects the run times of the algorithms in different problems settings.

In Appendix A we show that HAC does not only have the same complexity, but it also performs exactly the same number of consistency checks as GAC-2001 in arc consistent problems. We also show that in arc inconsistent problems there can be a difference in the number of checks in favor of the HVE.

### 3.2 Search Algorithms

Search algorithms that maintain local consistencies are widely used for CSP solving. Some of them have been extended to the non-binary case. For example, maintaining arc consistency (MAC) and forward checking (FC). It has been shown that the non-binary version of MAC (MGAC) applied in a non-binary CSP is equivalent to MAC applied in the HVE of the CSP when only original variables are instantiated and the same variable orderings are used (Stergiou & Walsh, 1999). We show that, like MGAC, non-binary extensions of FC can be emulated by equivalent algorithms that run in the HVE.

FC (Haralick & Elliot, 1980) was first generalized to handle non-binary constraints by Van Hentenryck (1989). According to the definition of Van Hentenryck (1989), forward checking is performed after the $k$-1 variables of an $k$-ary constraint have been assigned and the remaining variable is unassigned. This algorithm is called nFC0 in the paper by Bessiére, Meseguer, Freuder, & Larrosa (2002) where more, and stronger, generalizations of FC to non-binary constraints were introduced. These generalizations differ between them in the extent of look-ahead they perform after each variable instantiation. Algorithm nFC1 applies one pass of GAC on each constraint or constraint projection involving the current variable and exactly one future variable[4]. Algorithm nFC2 applies GAC on the set of constraints involving the current variable and at least one future variable, in one pass. Algorithm nFC3 applies GAC to the set of constraints involving the current variable and at least one future variable. Algorithm nFC4 applies GAC on the set of constraints involving at least one past variable and at least one future variable, in one pass. Algorithm nFC5, which is the strongest version, applies GAC to the set of constraints involving at least one past variable and at least one future variable. All the generalizations reduce to simple FC when applied to binary constraints.

---
4. "One pass" means that each constraint is processed only once.

We will show that the various versions of nFC are equivalent, in terms of visited nodes, to binary versions of FC that run in the HVE of the problem. This holds under the assumption that the binary algorithms only assign original variables and they use the same variable and value ordering heuristics, static or dynamic, as their non-binary counterparts. Note that if such an algorithm finds a consistent assignment for all original variables, and these assignments are propagated to the dual variables, then all the domains of the dual variables will be reduced to singletons. That is, the domain of each dual variable $v_c$ will only contain the single tuple that is consistent with the assignments of the original variables constrained with $v_c$. Therefore, the algorithm can proceed to assign the dual variables in a backtrack-free manner.

The equivalence between nFC1 and a version of FC for the HVE, called FC+ (Bacchus & van Beek, 1998), has been proved by Bessiére et al. (2002). FC+ is a specialized forward checking algorithm for the HVE. It operates like standard binary FC except that when the domain of a dual variable is pruned, FC+ removes from adjacent original variables any value which is no longer supported.

Algorithms nFC2-nFC5 also have equivalent algorithms that operate in the HVE. We call these algorithms hFC2–hFC5. For example, hFC5 will enforce AC on the set of dual variables, and original variables connected to them, such that each dual variable is connected to at least one past original variable and at least one future original variable. Note that nFC0 has no natural equivalent algorithm in the HVE. If we emulate it in the HVE we will get an inefficient and awkward algorithm. In the following, hFC0 will refer to the standard binary FC algorithm and hFC1 will refer to FC+.

**Proposition 3.1** In any non-binary CSP, under a fixed variable and value ordering, algorithm nFCi, i= 2, . . . 5, is equivalent to algorithm hFCi that operates in the hidden variable encoding of the problem.

**Proof:** We prove this for nFC5, the strongest among the generalized FC algorithms. The proof for the other versions is similar. We only need to prove that at each node of the search tree algorithms nFC5 and hFC5 will delete exactly the same values from the domains of original variables. Assume that at some node, after instantiating the current variable, nFC5 deletes value $a$ from a future variable $x_i$. Then there exists some constraint $c$ including $x_i$ and at least one past variable, and value $a$ of $x_i$ has no supporting tuple in $c$. In the HVE, when hFC5 tries to make $v_c$ (the dual variable corresponding to $c$) AC it will remove all tuples that assign $a$ to $x_i$. Hence, hFC5 will delete $a$ from the domain of $x_i$. Now in the opposite case, if hFC5 deletes value $a$ from an original variable $x_i$ it means that all tuples including that assignment will be removed from the domains of dual variables that include $x_i$ and at least one past variable. In the non-binary representation of the problem, the assignment of $a$ to $x_i$ will not have any supporting tuples in constraints that involve $x_i$ and at least one past variable. Therefore, nFC5 will delete $a$ from the domain of $x_i$. □

Algorithms nFC2–nFC5 are not only equivalent in node visits with the corresponding algorithms hFC2–hFC5, but they also have the same asymptotic cost. This holds under the condition that the non-binary algorithms use GAC-2001 (or some other optimal algorithm) to enforce GAC, and the HVE versions use algorithm HAC.

**Proposition 3.2** In any non-binary CSP, under a fixed variable and value ordering, algorithm nFCi, i= 2, . . . 5, has the same asymptotic cost as algorithm hFCi that operates in the hidden variable encoding of the problem.

**Proof:** In Section 3.1 we showed that we can enforce AC on the HVE of a non-binary CSP with the same worst-case complexity as GAC in the non-binary representation of the problem. Since algorithm nFCi enforces GAC on the same part of the problem on which algorithm hFCi enforces AC, and they visit the same nodes of the search tree, it follows that the two algorithm have the same asymptotic cost. □

In the paper by Bessiére et al. (2002), a detailed discussion on the complexities of algorithms nFC0–nFC5 is made. The worst-case complexity of nFC2 and nFC3 in one node is $O(|C_{c,f}|(k-1)d^{k-1})$, where $|C_{c,f}|$ is the number of constraints involving the current variable and at least one future variable. This is also the complexity of hFC3 and hFC4. The worst-case complexity of nFC4 and nFC5 in one node is $O(|C_{p,f}|(k-1)d^{k-1})$, where $|C_{p,f}|$ is the number of constraints involving at least one past variable and at least one future variable. This is also the complexity of hFC4 and hFC5.

Assuming that $nodes(alg_i)$ is the set of search tree nodes visited by search algorithm $alg_i$ then the following holds.

**Corollary 3.1** Given the hidden variable encoding of a CSP with fixed variable and value ordering schemes, the following relations hold:

1. nodes(hFC1)⊆ nodes(hFC0)

2. nodes(hFC2)⊆ nodes(hFC1)

3. nodes(hFC5)⊆ nodes(hFC3)⊆ nodes(hFC2)

4. nodes(hFC5)⊆ nodes(hFC4)⊆ nodes(hFC2)

5. nodes(MAC)⊆ nodes(hFC5)

**Proof:** The proof of 1 is straightforward, see the paper by Bacchus & van Beek (1998). Proof of 2-4 is a straightforward consequence of Proposition 3.1 and Corollary 2 from the paper by Bessiére et al. (2002) where the hierarchy of algorithms nFC0-nFC5 in node visits is given. It is easy to see that 5 holds since hFC5 applies AC in only a part of the CSP, while MAC applies it in the whole problem. Therefore, MAC will prune at least as many values as hFC5 at any given node of the search tree. Since the same variable and value ordering heuristics are used, this means that MAC will visit at most the same number of nodes as hFC5. □

Note that in the paper by Bacchus & van Beek (1998) experimental results show differences between FC in the HVE and FC in the non-binary representation. However, the algorithms compared there were FC+ and nFC0, which are not equivalent. Also, it has been proved that hFC0 can have an exponentially greater cost than nFC0, and vice versa (Bacchus et al., 2002). However, these algorithms are not equivalent. As proved in Proposition 3.2, the result of Bacchus et al. (2002) does not hold when comparing equivalent algorithms.

So far we have showed that solving a non-binary CSP directly is in many ways equivalent to solving it using the HVE, assuming that only original variables are instantiated. A natural question is whether there are any search techniques which are inapplicable in the non-binary case, but can be applied in the encoding. The answer is the ability of a search algorithm that operates on the encoding to instantiate dual variables. In the equivalent non-binary representation this would imply instantiating values of more than one variables simultaneously. To implement such an algorithm we would have to modify standard search algorithms and heuristics or devise new ones. On the other hand, in the HVE an algorithm that instantiates dual variables can be easily implemented.

## 4. Algorithms for the Dual Encoding

In this section we turn our attention to the DE and describe a specialized AC algorithm with significantly lower complexity than a generic algorithm.

### 4.1 Arc Consistency

We know that AC in the DE is strictly stronger than GAC in the non-binary representation and AC in the HVE (Stergiou & Walsh, 1999). Since the DE is a binary CSP, one obvious way to apply AC is using a generic AC algorithm. The domain size of a dual variable corresponding to a $k-$ary constraint is $d^k$ in the worst case. Therefore, if we apply an optimal AC algorithm then we can enforce AC on one dual constraint with $O(d^{2k})$ worst-case complexity. In the DE of a CSP with $e$ constraints of maximum arity $k$ there are at most $e(e-1)/2$ binary constraints (when all pairs of dual variables share one or more original variables). Therefore, we can enforce AC in the DE of the CSP with $O(e^2 d^{2k})$ worst-case complexity. This is significantly more expensive compared to the $O(ekd^k)$ complexity bound of GAC in the non-binary representation and AC in the HVE. Because of the very high complexity bound, AC processing in the DE is considered to be impractical, except perhaps for very tight constraints.

However, we will now show that AC can be applied in the DE much more efficiently. To be precise we can enforce AC on the DE of a non-binary CSP with $O(e^3 d^k)$ worst-case time complexity. The improvement in the asymptotic complexity can be achieved by exploiting the structure of the DE; namely, the fact that the constraints in the DE are piecewise functional.

Consider a binary constraint between dual variables $v_i$ and $v_j$. We can create a piecewise decomposition of the tuples in the domain of either dual variable into groups such that all tuples in a group are supported by the same group of tuples in the other variable. If the non-binary constraints corresponding to the two dual variables share $f$ original variables $x_1, \ldots, x_f$ of domain size $d$, then we can partition the tuples of $v_i$ and $v_j$ into $d^f$ groups. Each tuple in a group $s$ includes the same sub-tuple of the form $(a_1, \ldots, a_f)$, where $a_1 \in D(x_1), \ldots, a_f \in D(x_f)$. Each tuple $\tau$ in $s$ will be supported by all tuples in a group $s'$ of the other variable, where each tuple in $s'$ also includes the sub-tuple $(a_1, \ldots, a_f)$. The tuples belonging to $s'$ will be the only supports of tuple $\tau$ since any other tuple does not contain the sub-tuple $(a_1, \ldots, a_f)$. In other words, a group of tuples $s$ in variable $v_i$ will only be supported by a corresponding group $s'$ in variable $v_j$ where the tuples in both groups have the same values for the original variables that are common to the two encoded non-binary constraints. Therefore, the constraints in the DE are piecewise functional.

**Example 4.1** Assume that we have two dual variables $v_1$ and $v_2$. $v_1$ encodes constraint $(x_1, x_2, x_3)$, and $v_2$ encodes constraint $(x_1, x_4, x_5)$, where the original variables $x_1, \ldots, x_5$ have the domain $\{0, 1, 2\}$. We can partition the tuples in each dual variable into 3 groups. The first group will include tuples of the form $(0, *, *)$, the second will include tuples of the form $(1, *, *)$, and the third will include tuples of the form $(2, *, *)$. A star $(*)$ means that the corresponding original variable can take any value. Each group is supported only by the corresponding group in the other variable. Note that the tuples of a variable $v_i$ are partitioned in different groups according to each constraint that involves $v_i$. For instance, if there is another dual variable $v_3$ encoding constraint $(x_6, x_7, x_3)$ then the partition of tuples in $D(v_1)$ according to the constraint between $v_1$ and $v_3$ is into groups of the form $(*, *, 0), (*, *, 1), (*, *, 2)$.

Van Hentenryck, Deville & Teng (1992) have shown that AC can be achieved in a set of binary piecewise functional constraints with $O(ed)$ worst-case time complexity, an improvement of $O(d)$ compared to the $O(ed^2)$ complexity of arbitrary binary constraints (Van Hentenryck et al., 1992). Since we showed that the constraints in the DE are piecewise functional, the result of Van Hentenryck et al. (1992) means that we can improve on the $O(e^2 d^{2k})$ complexity of AC in the DE.

In Figure 4 we sketch an AC-3 like AC algorithm specifically designed for the DE, which we call PW-AC (*PieceWise Arc Consistency*). As we will show, this algorithm has a worst-case time complexity of $O(e^3 d^k)$. The same complexity bound can be achieved by the AC-5 algorithm of Van Hentenryck et al. (1992), in its specialization to piecewise functional constraints, with the necessary adaptations to operate in the DE. As do most AC algorithms, PW-AC uses a stack (or queue) to propagate deletions from the domains of variables. This stack processes groups of piecewise decompositions, instead of variables or constraints as is usual in AC algorithms. We use the following notation:

- $S(v_i, v_j) = \{s_1(v_i, v_j), \ldots, s_m(v_i, v_j)\}$ denotes the piecewise decomposition of $D(v_i)$ with respect to the constraint between $v_i$ and $v_j$. Each $s_l(v_i, v_j)$, $l = 1, \ldots, m$, is a group of the partition.

- $sup(s_l(v_i, v_j))$ denotes the group of $S(v_j, v_i)$ that can support group $s_l(v_i, v_j)$ of $S(v_i, v_j)$. As discussed, this group is unique.

- $counter(s_l(v_i, v_j))$ holds the number of valid tuples that belong to group $s_l(v_i, v_j)$ of decomposition $S(v_i, v_j)$. That is, at any time the value of $counter(s_l(v_i, v_j))$ gives the current cardinality of the group.

- $GroupOf(S(v_i, v_j), \tau)$ is a function that returns the group of $S(v_i, v_j)$ where tuple $\tau$ belongs. To implement this function, for each constraint between dual variables $v_i$

**function** $PW - AC$
1:    $Q \leftarrow \emptyset$
2:    initialize all group counters to 0
3:    **for** each variable $v_i$
4:        **for** each variable $v_j$ constrained with $v_i$
5:            **for** each tuple $\tau \in D(v_i)$
6:                $counter(GroupOf(S(v_i, v_j), \tau)) \leftarrow counter(GroupOf(S(v_i, v_j), \tau)) + 1$
7:    **for** each variable $v_i$
8:        **for** each variable $v_j$ constrained with $v_i$
9:            **for** each group $s_l(v_i, v_j)$
10:               **if** $counter(s_l(v_i, v_j)) = 0$
11:                  put $s_l(v_i, v_j)$ in $Q$
12:    **return** $Propagation$

**function** $Propagation$
13:    **while** $Q$ is not empty
14:        pop group $s_l(v_i, v_j)$ from $Q$
15:            $\delta \leftarrow \emptyset$
16:            $\delta \leftarrow Revise(v_i, v_j, s_l(v_i, v_j))$
17:                **if** $D(v_j)$ is empty **return** INCONSISTENCY
18:                **for** each group $s_{l'}(v_j, v_k)$ in $\delta$ put $s_{l'}(v_j, v_k)$ in $Q$
19:    **return** CONSISTENCY

**function** $Revise(v_i, v_j, s_l(v_i, v_j))$
20:    **for** each tuple $\tau \in D(v_j)$ where $\tau \in sup(s_l(v_i, v_j))$
21:        remove $\tau$ from $D(v_j)$
22:        **for** each group $s_{l'}(v_j, v_k)$ that includes $\tau$
23:            $counter(s_{l'}(v_j, v_k)) \leftarrow counter(s_{l'}(v_j, v_k)) - 1$
24:            **if** $counter(s_{l'}(v_j, v_k)) = 0$
25:                add $s_{l'}(v_j, v_k)$ to $\delta$
26:    **return** $\delta$

Figure 4: PW-AC. An AC algorithm for the dual encoding.

and $v_j$ we store the original variables shared by the non-binary constraints $c_{v_i}$ and $c_{v_j}$. Also, for each such original variable $x_l$ we store $pos(x_l, c_{v_i})$ and $pos(x_l, c_{v_j})$. In this way the $GroupOf$ function takes constant time.

- The set $\delta$ contains the groups that have their counter reduced to 0 after a call to function $Revise$. That is, groups such that all tuples belonging to them have been deleted.

The algorithm works as follows. In an initialization phase, for each group we count the number of tuples it contains (lines 3–6). Then, for each variable $v_i$ we iterate over the variables $v_j$ that are constrained with $v_i$. For each group $s_l(v_i, v_j)$ of $S(v_i, v_j)$, we check if $s_l(v_i, v_j)$ is empty or not (line 10). If it is empty, it is added to the stack for propagation.

In the next phase, function *Propagation* is called to delete unsupported tuples and propagate the deletions (line 12). Once the previous phase has finished, the stack will contain a number of groups with 0 cardinality. For each such group $s_l(v_i, v_j)$ we must remove all tuples belonging to group $sup(s_l(v_i, v_j))$ since they have lost their support. This is done by successively removing a group $s_l(v_i, v_j)$ from the stack and calling function *Revise*. Since group $sup(s_l(v_i, v_j))$ has lost its support, each tuple $\tau \in D(x_j)$ that belongs to $sup(s_l(v_i, v_j))$ is deleted (lines 20–21). Apart from $sup(s_l(v_i, v_j))$, tuple $\tau$ may also belong to other groups that $D(v_j)$ is partitioned in with respect to constraints between $v_j$ and other variables. Since $\tau$ is deleted, the counters of these groups must be updated (i.e. reduced by one). This is done in lines 22–23. In the implementation we use function *GroupOf* to access the relevant groups. If the counter of such a group becomes 0 then the group is added to the stack for propagation (lines 24–25 and 18). The process stops when either the stack or the domain of a variable becomes empty. In the former case, the DE is AC, while in the latter it is not.

The following example illustrates the advantage of algorithm PW-AC over both a generic AC algorithm employed in the DE, and AC in the HVE (or GAC in the non-binary representation).

**Example 4.2** Consider three constraints $c_1$, $c_2$, $c_3$ as part of a CSP, where $vars(c_1) = \{x_0, x_1, x_3\}$, $vars(c_2) = \{x_2, x_3, x_4\}$, $vars(c_3) = \{x_2, x_4, x_5\}$. Assume that at some point the domains of the variables in the DE of the problem are as shown in Figure 5 (disregarding the original variables depicted with dashed lines). Assume that we try to enforce AC in the

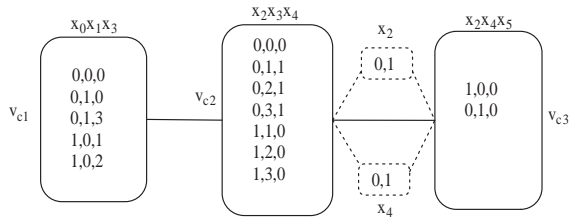

Figure 5: Dual encoding of a non-binary CSP.

DE using algorithm AC-2001[5]. The algorithm will discover that the first tuple in $D(v_{c_2})$ has no support in $D(v_{c_3})$ (there is no tuple with $x_2 = 0$ and $x_4 = 0$) and will delete it. Because of this deletion, the first two tuples in $D(v_{c_1})$ lose their support in $D(v_{c_2})$ and AC-2001 must therefore look for new supports. For each of the two tuples of $D(v_{c_1})$ the algorithm will check all the 6 remaining tuples in $D(v_{c_2})$ before discovering that there is no support. As a result the two tuples will be deleted. Algorithm PW-AC, on the other hand, will set the counter of the group where the first tuple of $D(v_{c_2})$ belongs (according to partition $S(v_{c_2}, v_{c_3})$) to 0 once it deletes the tuple. This will result in a call to function

---

5. Note that we can construct similar examples for any generic AC algorithm.

*Revise* and an automatic deletion of the first two tuples of $D(v_{c_1})$, saving a total of $2 \times 6$ checks.

Now consider the HVE of the problem. Applying AC on the HVE will have no effect because values 0 and 1 of $x_2$ and $x_4$ are both supported in $D(v_{c_2})$ and $D(v_{c_3})$. Therefore there is no propagation through these variables, and as a result the two tuples of $D(v_{c_1})$ will not be deleted. Similarly, there will be no propagation if we apply GAC in the non-binary representation of the problem.

Note that the theoretical results regarding the DE presented in the rest of the paper hold if the AC-5 algorithm of Van Hentenryck et al. (1992) was adapted and used the DE instead of PW-AC. The two algorithms have some similarities (e.g. they both use a function to access the group of a decomposition that a certain tuple belongs to, though implemented differently), but their basic operation is different. The algorithm of Van Hentenryck et al. (1992), being an instantiation of AC-5, handles a queue of triples $(v_i, v_j, a)$ to implement constraint propagation, where $v_i$ and $v_j$ are two variables involved in a constraint and $a$ is a value that has been removed from $D(v_j)$. PW-AC utilizes a queue of piecewise decompositions. Also the data structures used by the algorithms are different. PW-AC checks and updates counters to perform the propagation which, as we explain below, requires space exponential in the number of common variables in the non-binary constraints. The algorithm of Van Hentenryck et al. (1992) utilizes a more complicated data structure which requires space exponential in the arity of the non-binary constraints. It has to be noted, however, that PW-AC is specifically designed for the DE. That is, its operation, data structures, and the way it checks for consistency are based on the fact that the domains of the dual variables consist of the tuples of the original constraints extensionally stored. On the other hand, the algorithm of Van Hentenryck et al. (1992) is generic, in the sense that it can be adapted to operate on any piecewise functional constraint.

### 4.1.1 COMPLEXITIES

The PW-AC algorithm consists of two phases. In the initialization phase we set up the group counters, and in the main phase we delete unsupported tuples and propagate the deletions. We now analyze the time complexity of PW-AC. Note that in our complexity analysis we measure operations, such as incrementing or decrementing a counter, since PW-AC does not perform consistency checks in the usual sense.

**Proposition 4.1** The worst-case time complexity of algorithm PW-AC is $O(e^3 d^k)$.

**Proof:** We assume that for any constraint in the dual encoding, the non-binary constraints corresponding to the two dual variables $v_i$ and $v_j$ share at most $f$ original variables $x_1, \ldots, x_f$ of domain size $d$. This means that each piecewise decomposition consists of at most $d^f$ groups. Obviously, $f$ is equal to $k-1$, where $k$ is the maximum arity of the constraints. In the initialization phase of lines 3–6 we iterate over all constraints, and for each constraint between variables $v_i$ and $v_j$, we iterate over all the tuples in $D(v_i)$. This is done with $O(e^2 d^k)$ asymptotic time complexity. Then, all empty groups are inserted in $Q$ (lines 7–11). This requires $e^2 d^f$ operations in the worst case. After the initialization, function *Propagation* is called. A group is inserted to $Q$ (and later removed) only when it becomes empty. This means that the **while** loop of *Propagation* is executed at most $d^f$ times for each constraint, and at most $e^2 d^f$ times in total. This is also the maximum number of times function *Revise* is called (once in every iteration of the loop). The cost of function *Revise* is computed as follows: Assuming *Revise* is called for a group $s_l(v_i, v_j)$, we iterate over the (at most) $d^{k-f}$ tuples of group $sup(s_l(v_i, v_j))$ (line 20). In each iteration we remove a tuple $\tau$ (line 21) and we update the counters of the groups where $\tau$ belongs (lines 22–23). There are at most $e$ such groups (in case $v_j$ is constrained with all other dual variables). Therefore, each iteration costs $O(e)$, and as a result, each call to *Revise* costs $O(ed^{k-f})$. Since *Revise* is called at most $e^2 d^f$ times, the complexity of PW-AC, including the initialization step, is $O(e^2 d^k + e^2 d^f + e^2 d^f e d^{k-f}) = O(e^3 d^k)$. □

Note that PW-AC can be easily used incrementally during search. In this case, the initialization phase will only be executed once. The asymptotic space complexity of PW-AC, and any AC algorithm on a binary encoding, is dominated by the $O(ed^k)$ space need to store the allowed tuples of the non-binary constraints. Algorithm PW-AC also requires $O(e^2 d^f)$ space to store the counters for all the groups, $O(e^2 d^f)$ space for the stack, and $O(fe^2)$ space for the fast implementation of function $GroupOf$.

## 5. Algorithms for the Double Encoding

The double encoding has rarely been used in experiments with binary encodings, although it combines features of the HVE and the DE, and therefore may exploit the advantages of both worlds. To be precise, the double encoding offers the following interesting potential: search algorithms can deploy dynamic variable ordering heuristics to assign values to the original variables, while constraint propagation can be implemented through the constraints between dual variables to achieve higher pruning. In this section we first briefly discuss how AC can be applied in the double encoding. We then show how various search algorithms can be adapted to operate in the double encoding.

### 5.1 Arc Consistency

AC can be enforced on the double encoding using algorithm PW-AC with the addition that each time a value $a$ of an original variable $x_i$ loses all its supports in an adjacent dual variable, it is deleted from $D(x_i)$. Alternatively, we can use any generic AC algorithm, such as AC-2001. Note that an AC algorithm applied in the double encoding can enforce various levels of consistency depending on which constraints it uses for propagation between dual variables. That is, propagation can be either done directly through the constraints between dual variables, or indirectly through the constraints between dual and original variables. For example, if we only use the constraints between dual and original variables then we get the same level of consistency as AC in the HVE. If propagation between dual variables is performed using the constraints of the DE then we get the same level of consistency as AC in the DE, for the dual variables, and we also prune the domains of the original variables. In between, we have the option to use different constraints for propagation in different parts of the problem. As the next example shows, AC in the double encoding can achieve a very high level of consistency compared to the non-binary representation. In Sections 6.2 and 6.3 we will show that this can have a profound effect in practice.

**Example 5.1** Consider the problem of Figure 6. Applying AC through the constraint between the two dual variables will determine that the problem is insoluble. However, the problem in its non-binary representation is not only GAC, but also singleton (generalized) arc consistent (SGAC), which is a very high level of consistency. A CSP is SGAC if after applying GAC in the problem induced by any instantiation of a single variable, there is no domain wipeout (Debruyne & Bessière, 2001; Prosser, Stergiou, & Walsh, 2000).

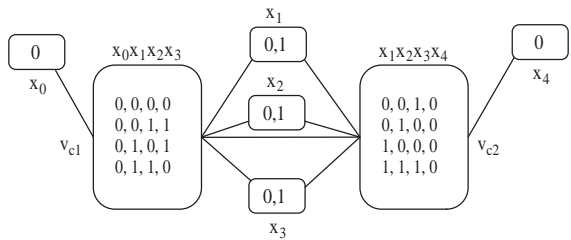

Figure 6: Double encoding of a problem that is not AC in the double encoding but is SGAC in the non-binary representation.

### 5.2 Search Algorithms

Various search algorithms for the double encoding can be defined, depending on the variables that are instantiated and the constraints that are used for propagation. Here we will restrict ourselves to algorithms that only instantiate original variables and perform propagation using the constraints between dual variables. Intuitively this is the most interesting class of algorithms because they combine nice features from the non-binary representation and the HVE (small domain sizes), and from the DE (strong propagation).

We first show that the FC versions for the HVE discussed in Section 3.2 can be adapted to yield algorithms that run on the double encoding. We call these algorithms dFC0–dFC5. Each algorithm dFC$i$ ($i = 0, \ldots, 5$) instantiates only original variables and enforces AC on exactly the same set of variables of the double encoding as the corresponding algorithm hFC$i$ does in the HVE. For example, dFC5 will enforce AC on the set of dual variables, and original variables connected to them, such that each dual variable is connected to at least one past original variable and at least one future original variable. The difference between algorithm dFC$i$ ($i = 2, \ldots, 5$) and hFC$i$ is that the former can exploit the constraints between dual variables to enforce a higher level of consistency than the latter. Not surprisingly, this results in stronger algorithms.

**Proposition 5.1** In any non-binary CSP, under a fixed variable and value ordering, algorithm dFCi (i= 2, ..., 5) is strictly stronger than the respective algorithm hFCi.

**Proof:** It is easy to see that if a value is pruned by hFCi in the HVE then it is also pruned by dFCi in the double encoding. This is a straightforward consequence of the fact that 1) the double encoding subsumes the HVE, and 2) algorithms dFCi and hFCi enforce AC on the same set of variables. Algorithm dFCi is strictly stronger than hFCi because, by exploiting the constraints between dual variables, it can prune more values than hFCi. Consider, for instance, a problem with two constraints $c_1$ and $c_2$, where $vars(c_1) = \{x_1, x_2, x_3, x_4\}$ and $vars(c_2) = \{x_1, x_2, x_3, x_5\}$. All variables $x_i$, $i = 1, \ldots, 5$, have domains $\{0, 1\}$. The allowed tuples of the constraints are $rel(c_1) = \{(0, 0, 1, 0), (0, 1, 0, 1), (1, 1, 0, 1)\}$ and $rel(c_2) = \{(0, 0, 0, 0), (0, 1, 1, 1), (1, 0, 0, 0)\}$. If $x_1$ is given value 0 in the HVE then algorithms hFC2–hFC5 will prune tuples $(1, 1, 0, 1)$ and $(1, 0, 0, 0)$ from the domains of dual variables $v_{c_1}$ and $v_{c_2}$ respectively. No other pruning will be performed. In the double encoding, the same variable assignment, by any of the algorithms dFC2–dFC5, will cause the domain wipe-out of the two dual variables. $\square$

**Corollary 5.1** In any non-binary CSP, under a fixed variable and value ordering, algorithm dFCi (i= 2 . . . 5) is strictly stronger than the respective algorithm nFCi (i=2 . . . 5).
   **Proof:** Straightforward consequence of Propositions 5.1 and 3.1. $\square$

It is easy to see that algorithm hFC0 (i.e. simple binary FC) is equivalent to dFC0. The same holds for algorithms hFC1 and dFC1. As with the various versions of FC, the MAC algorithm can be adapted to run in the the double encoding so that only original variables are instantiated, and propagation is implemented through the constraints between dual variables. It is easy to see that this algorithm is strictly stronger than the corresponding algorithm for the HVE (the proof is similar to the proof of Proposition 5.1). More interestingly, we show that this MAC algorithm for the double encoding can, at most, have a polynomially greater cost than the corresponding MAC algorithm for the HVE, while, on the other hand, it can be exponentially better.

**Proposition 5.2** In any non-binary CSP, under a fixed variable and value ordering, the MAC algorithm for the hidden variable encoding that only instantiates original variables can have exponentially greater cost than the corresponding MAC algorithm for the double encoding.
   **Proof:** To prove this, we can use Example 14 from the paper by Bacchus et al. (2002). In this example we have a CSP with $4n + 2$ variables, $x_1, \ldots, x_{4n+2}$, each with domain $\{1, \ldots, n\}$, and $2n + 1$ constraints:

$c_1$: $(x_1 + x_2 \bmod 2) \neq (x_3 + x_4 \bmod 2)$

$c_2$: $(x_3 + x_4 \bmod 2) \neq (x_5 + x_6 \bmod 2)$

. . .

$c_{2n}$: $(x_{4n-1} + x_{4n} \bmod 2) \neq (x_{4n+1} + x_{4n+2} \bmod 2)$

$c_{2n+1}$: $(x_{4n+1} + x_{4n+2} \bmod 2) \neq (x_1 + x_2 \bmod 2)$

Assume that the variables are assigned in lexicographic order in the double encoding. If $x_1$ and $x_2$ are assigned values such that $(x_1 + x_2 \bmod 2) = 0$ then enforcing AC will prune any tuples from $D(v_{c_1})$ such that $(x_3 + x_4 \bmod 2) = 0$. This in turn will prune from $D(v_{c_2})$ any tuples such that $(x_5 + x_6 \bmod 2) = 1$. Continuing this way, AC propagation will prune from $D(v_{c_{2n+1}})$ any values such that $(x_1 + x_2 \bmod 2) = 0$. When these deletions are propagated to $v_{c_1}$, $D(v_{c_1})$ will become empty. In a similar way, enforcing AC after assignments to $x_1$ and $x_2$, such that $(x_1 + x_2 \ mod \ 2) = 1$, leaves $D(v_{c_1})$ empty. Therefore, the CSP is insoluble. MAC in the double encoding needs to instantiate two variables to discover this, and visit $O(n^2)$ nodes. On the other hand, as explained by Bacchus et al. (2002), MAC in the HVE needs to visit $O(n^{log(n)})$ nodes to conclude that the problem is insoluble. Finally, note that, for each node, the asymptotic costs of MAC in the double encoding (using PW-AC) and MAC in the HVE are polynomially related. Therefore, MAC in the HVE can be exponentially worse than MAC in the double encoding. □

A corollary of Proposition 5.2 is that MAC in the double encoding can have have exponentially smaller cost than MGAC in the non-binary representation.

**Proposition 5.3** In any non-binary CSP, under a fixed variable and value ordering, the MAC algorithm for the double encoding that only instantiates original variables can have at most polynomially greater cost than the corresponding MAC algorithm for the hidden variable encoding.

**Proof:** To prove this we need to show two things: 1) The number of node visits made by MAC in the double encoding is at most by a polynomial factor greater than the number of node visits made by MAC in the HVE, 2) At each node, the worst-case cost of MAC in the double encoding is at most by a polynomial factor greater than the worst-case cost of AC in the HVE. The former is true since MAC in the double encoding is strictly stronger than MAC in the HVE. The latter can be established by considering the worst case complexities of the algorithms at each node. MAC in the HVE costs $O(ekd^k)$ at each node, while MAC in the double encoding can use PW-AC to enforce AC, which costs $O(e^3d^k)$. Therefore, there is only a polynomial difference. □

In a similar way, we can prove that the relationship of Proposition 5.3 holds between each algorithm dFCi (i= 2...5) and the corresponding algorithm hFCi. A corollary of Proposition 5.3 is that MAC in the double encoding can have at most polynomially greater cost than MGAC in the non-binary representation. It is important to note that Proposition 5.3 holds only if algorithm PW-AC is used to enforce AC in the double encoding. If we use a generic algorithm, like AC-2001, then we can get exponential differences in favor of MAC in the HVE. Finally, regarding the relationship in node visits among the algorithms for the double encoding, we have the following.

**Proposition 5.4** Given the double encoding of a CSP with fixed variable and value ordering schemes, the following relations hold:

1. nodes(dFC1)⊆ nodes(dFC0)

2. nodes(dFC2)⊆ nodes(dFC1)

3. nodes(dFC5)⊆ nodes(dFC3)⊆ nodes(dFC2)

4. nodes(dFC5)⊆ nodes(dFC4)⊆ nodes(dFC2)

5. nodes(dMAC)⊆ nodes(dFC5)

**Proof:** The proof is very simple and it is based on comparing the size of the subsets of the problem where each algorithm enforces AC. □

## 6. Experimental Results

In this section we make an empirical study of algorithms for binary encodings. The empirical study is organized in two parts:

- In the first part (Subsections 6.1 and 6.2) we evaluate the improvements offered by specialized algorithms compared to generic ones. At the same time we compare the efficiency of algorithms that run in the binary encodings to their non-binary counterparts. This comparison can give us a better understanding of when encoding a non-binary problem into a binary one pays off, and which encoding is preferable. For this empirical investigation we use randomly generated problems, random problems with added structure, and benchmark crossword puzzle generation problems. Random problems allow us to relate the performance of the algorithms to certain parameters, such as tightness, constraint graph density, and domain size. Crossword puzzles are standard benchmarks for comparing binary to non-binary constraint models, and allow us to to evaluate the performance of the algorithms on problems that include constraints of high arity.

- In the second part (Subsection 6.3) , we investigate the usefulness of binary encodings in more realistic problem settings. For this study we use problems from the domains of configuration and frequency assignment and we compare the performance of MAC algorithms that run in the encodings to an MGAC algorithm in the non-binary representation.

All algorithms were implemented in C. All experiments were run on a PC with a 3.06 GHz Pentium 4 processor and 1 GB RAM. In all experiments, all algorithms use the dom/deg heuristic (Bessière & Régin, 1996b) for dynamic variable ordering and lexicographic value ordering.

### 6.0.1 Random Problems

Random instances were generated using the extended *model B* as it is described by Bessiére et al. (2002). To summarize this generation method, a random non-binary CSP is defined by the following five input parameters:

**n** - number of variables

**d** - uniform domain size

**k** - uniform arity of the constraints

**p** - density (%) percentage of the generated graph, i.e. the ratio between existing constraints and the number of possible sets of $k$ variables

**q** - uniform looseness (%) percentage of the constraints, i.e. the ratio between allowed tuples and the $d^k$ total tuples of a constraint

The constraints and the allowed tuples were generated following a uniform distribution. We made sure that the generated graphs were connected. In the following, a class of non-binary

CSPs will be denoted by a tuple of the form $<n,d,k,p,q>$. We use a star ($*$) for the case where one of the parameters is varied. For example, the tuple $<50,20,3,10,*>$ stands for the class of problems with 50 variables, domain size 20, arity 3, graph density 10%, and varying constraint looseness.

### 6.0.2 Crossword Puzzles

Crossword puzzle generation problems have been used for the evaluation of algorithms and heuristics for CSPs (Ginsberg, Frank, Halpin, & Torrance, 1990; Beacham, Chen, Sillito, & van Beek, 2001) and binary encodings of non-binary problems (Bacchus & van Beek, 1998; Stergiou & Walsh, 1999). In crossword puzzle generation we try to construct puzzles for a given number of words and a given grid which has to be filled in with words. This problem can be represented as either a non-binary or a binary CSP in a straightforward way. In the non-binary representation there is a variable for each letter to be filled in and a non-binary constraint for each set of $k$ variables that form a word in the puzzle. The domain of each variable consists of the low case letters of the English alphabet giving a domain size of 26. The allowed tuples of such a constraint are all the words with $k$ letters in the dictionary used. These are very few compared to the $26^k$ possible combinations of letters, which means that the constraints are very tight. In the DE there is a variable for each word of length $k$ in the puzzle and the possible values of such a variable are all the words with $k$ letters in the dictionary. This gives variables with large domains (up to 4072 values for the Unix dictionary that we used in the experiments). There are binary constraints between variables that intersect (i.e. they have a common letter). In the HVE there are all the original variables as well a set of dual variables, one for each non-binary constraint.

## 6.1 Hidden Variable Encoding

In our first empirical study we investigated the performance of two MAC algorithms that operate in the HVE, and compared them to MGAC-2001, their counterpart in the non-binary representation. The two MAC algorithms for the HVE are MHAC-2001, which stands for MAC in the HVE that only instantiates original variables, and MHAC-2001-$full$ which is a MAC algorithm that may instantiate any variable (dual or original) according to the heuristic choice. As stated by their names, all three algorithms use AC-2001 (GAC-2001) to enforce AC. Although we also run experiments with the various versions of FC, we do not include any results since these algorithms are inefficient on hard problems (especially on hard crossword puzzles). However, the qualitative comparison between FC-based algorithms for the HVE and the non-binary representation is similar to the comparison regarding MAC-based algorithms.

### 6.1.1 Random Problems

Table 1 shows the performance, measured in cpu time, of the algorithms on five classes of randomly generated CSPs. All classes are from the hard phase transition region. Classes 1, 2, 3, and 4 are sparse, while 5 is more dense. We do not include results on MHAC-2001-$full$ because experiments showed that this algorithm has very similar behavior to MHAC-2001. The reason is that, because of the nature of the constraints, the dom/deg heuristic almost always selects original variables for instantiation. In the rare cases where the heuristic selected dual variables, this resulted in a large increase in cpu time.

| class | MGAC-2001 | MHAC-2001 |
|---|---|---|
| 1: $<30, 6, 3, 1.847, 50>$ | 2.08 | **1.90** |
| 2: $<75, 5, 3, 0.177, 41>$ | 4.09 | **3.41** |
| 3: $<50, 20, 3, 0.3, 5>$ | 64.15 | **28.10** |
| 4: $<50, 10, 5, 0.001, 0.5>$ | 74.72 | **22.53** |
| 5: $<20, 10, 3, 5, 40>$ | **5.75** | 8.15 |

Table 1: Comparison of algorithms MGAC-2001 and MHAC-2001 on random classes of problems. Classes 1 and 2 taken from the paper by Bessiére et al. (2002). We give the average run times (in seconds) for 100 instances at each class. The "winning" time for each instance is given in bold. We follow this in the rest of the paper.

From Table 1 we can see that MHAC-2001 performs better than MGAC-2001 on the sparse problems. In general, for all the 3-ary classes we tried with density less than $3\% - 4\%$ the relative run time performance of MHAC-2001 compared to MGAC-2001 ranged from being equal to being around 2-3 times faster. In the very sparse class 4, which includes problems with 5-ary constraints, MHAC-2001 is considerably more efficient than MGAC-2001. This is due to the fact that for sparse problems with relatively large domain sizes the hard region is located at low constraint looseness (i.e. small domains for dual variables) where only a few operations are required for the revision of dual variables. Another factor contributing to the dominance of the binary algorithm in class 4 is the larger arity of the constraints. The non-binary algorithm requires more operations to check the validity of tuples when the tuples are of large arity, as explained in Section 3.1.

When the density of the graph increases (class 5), the overhead of revising the domains of dual variables and restoring them after failed instantiations slows down MHAC-2001, and as a result it is outperformed by MGAC-2001. For denser classes than the ones reported, the phase transition region is at a point where more than half of the tuples are allowed, and in such cases the non-binary algorithm performs even better.

### 6.1.2 Crossword Puzzles

Table 2 demonstrates the performance of search algorithms on various crossword puzzles. We used benchmark puzzles from the papers by Ginsberg et al. (1990) and Beacham et al. (2001). Four puzzles (15.06, 15.10, 19.03, 19.04) could not be solved by any of the algorithms within 2 hours of cpu time. Also, two puzzles (19.05 and 19.10) were arc inconsistent. In both cases GAC discovered the inconsistency slower than AC in HVE (around 3:1 time difference in 19.05 and 10:1 in 19.10) because the latter method discovered early the domain wipe-out of a dual variable.

In the rest of the puzzles we can observe that MHAC-2001 performs better than MGAC-2001 on the hard instances. For the hard insoluble puzzles MHAC-2001 is up to 3 times faster than MGAC-2001. This is mainly due to the large arity of the constraints in these classes[6]. Another interesting observation is that there can be significant differences between the performance of methods that may instantiate dual variables and those which instantiate only original ones. In many cases MAC-2001-$full$ managed to find a (different) solution than MHAC-2001 and MGAC-2001 earlier. On the other hand, MAC-2001-$full$ was subject to thrashing in some instances where other methods terminate. The fact that in all insoluble puzzles MAC-2001-$full$ did not do better than MHAC-2001 shows that its performance is largely dependent on the variable ordering scheme. In many cases MAC-2001-$full$ visited less nodes than MHAC-2001. However, this was not reflected to similar time performance difference because when a dual variable is instantiated MAC-2001-$full$ does more work than when an original one is instantiated. It has to instantiate automatically each original variable $x_i$ constrained with the dual variable and propagate these changes to other dual variables containing $x_i$.

| puzzle | $n$ | $e$ | MGAC-2001 | MHAC-2001 | MHAC-2001-$full$ |
|---|---|---|---|---|---|
| 15.02 | 80 | 191 | 3.70 | **3.58** | — |
| 15.04* | 76 | 193 | 40.84 | 38.88 | **29.75** |
| 15.07 | 74 | 193 | 94.65 | **46.36** | 44m |
| 19.02 | 118 | 296 | **34.62** | 35.08 | — |
| 19.08 | 134 | 291 | — | — | **5.45** |
| 6×6 | 12 | 36 | 9.39 | **5.07** | 5.49 |
| 7×7* | 14 | 49 | 14m | **8m** | 12m |
| 8×8* | 16 | 64 | 6m | **2m** | 5m |
| 10×10* | 20 | 100 | **11.81** | 11.85 | 15.39 |

Table 2: Comparison (in cpu time) between algorithms for the HVE and algorithms for the non-binary representation of crossword puzzles. $n$ is the number of words and $e$ is the number of blanks. All times are in seconds except those followed by "m" (minutes). A dash (—) is placed wherever the algorithm did not manage to find a solution within 2 hours of cpu time. Problems marked by (*) are insoluble. We only include problems that were reasonably hard for at least one algorithm and at the same time were solvable within 2 hours by at least one algorithm.

### 6.2 Dual and Double Encodings

In this empirical study we investigated the performance of algorithms for the DE and double encoding. We tried to answer the following three questions: 1) How efficient are specialized algorithms compared to generic algorithms? 2) Does the use of a specialized algorithm make the DE an effective option for solving non-binary CSPs? 3) Can we take advantage of the theoretical properties of the double encoding in practice? To answer these questions, we run experiments with random and structured problems to evaluate the benefits offered by the specialized algorithm PW-AC when maintaining AC during search. We compared the performance of two MAC algorithms; one that uses AC-2001 to enforce AC (MAC-2001), and another that uses PW-AC to enforce AC (MAC-PW-AC). We also compared these algorithms to an algorithm that maintains GAC in the non-binary representation using GAC-2001 (MGAC-2001), and to MAC algorithms that maintain AC in the double encoding using PW-AC (algorithm MAC-PW-ACd) and AC-2001 (algorithm MAC-2001d).

---

6. Puzzles 6×6-10×10 correspond to square grids with no blank squares.

### 6.2.1 Random Problems

We first give some indicative results of a comparison between the various algorithms using random problems. Figure 7 compares the time that is required to enforce AC in the DE and GAC in the non-binary representation, while Figures 8-10 compare algorithms that maintain these consistencies.

Figure 7 shows the average cpu times (in msecs) that PW-AC and AC-2001 take to enforce AC in the DE of 100 random CSPs with 50 variables with domain size 30, ternary constraints, and 0.3 graph density (58 constraints). We also include the average time GAC-2001 takes to enforce GAC on the non-binary representation of the generated instances. The looseness of the constraints is varied starting from a point where all instances are not GAC up until GAC and AC in the DE do not delete any values. There is significant difference in the performance of PW-AC compared to AC-2001 which constantly rises as the looseness of the constraints becomes higher. This is expected, since as the number of allowed tuples in a constraint grows, AC-2001 takes more and more time to find supports. GAC-2001 is faster than PW-AC (up to one order of magnitude) when the looseness in low, but the difference becomes smaller as the looseness grows.

Figure 8 shows cpu times for a relatively sparse class of problems with 30 variables and 40 ternary constraints ($p = 1$). Figure 9 shows cpu times and node visits for a denser class with 30 variables and 203 ternary constraints ($p = 5$). Along the x-axis we vary the domain size of the variables. All data points show average cpu times (in secs) over 100 instances taken from the hard phase transition region.

We can make the following observations: 1) MAC-PW-AC and MAC-PW-ACd are significantly faster (one order of magnitude) than MAC-2001 and MAC-2001d, respectively, in both classes of problems. 2) For both these classes, the non-binary representation is preferable to the DE (MGAC-2001 is two orders of magnitude faster in the denser class). In the sparser class, MAC in the double encoding (i.e. algorithm MAC-PW-ACd) is competitive with MGAC-2001 for small domain sizes, and considerably faster for larger domain sizes. The effect of the domain size in the relative performance of the algorithms is mainly due to a run time advantage of PW-AC compared to GAC-2001, and not to the higher consistency level achieved in the double encoding[7]. The run time advantage of PW-AC can be explained considering that, as the domain size increases, GAC-2001 has to check an increasing number of tuples for supports; an operation more costly than the counter updates of PW-AC. In the denser class MGAC-2001 is constantly faster than all the other algorithms for all domain sizes. This is not surprising considering the $O(e^3 d^k)$ and $O(ekd^k)$ complexities of PW-AC and GAC-2001 (i.e. factor $e$ becomes more significant).

In Figure 10 we compare algorithms MGAC-2001 and MAC-PW-ACd (the faster among the algorithms for the encodings) in a class of problems with 20 variables and 48 4-ary constraints. The other algorithms for the encodings were not competitive in this class of problems. We can see that MGAC-2001 is more efficient for small domain sizes, while for larger domain sizes MAC-PW-ACd can be up to one order of magnitude faster. However, for denser classes of problems these results are reversed.

---

7. This was verified by looking at the node visits of the two algorithms.

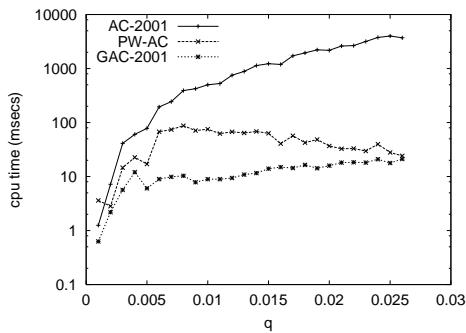

Figure 7: $<50, 30, 3, 0.3, *>$ CSPs.

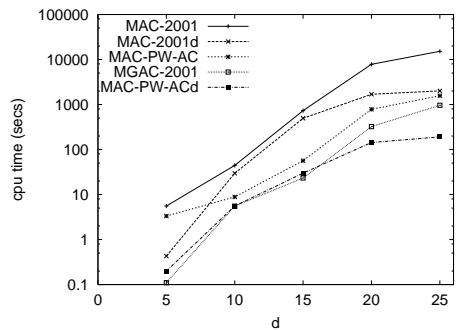

Figure 8: $<30, *, 3, 1, *>$ CSPs.

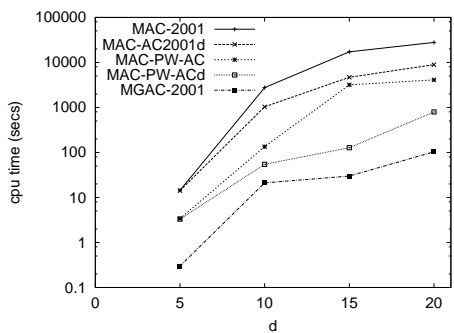

Figure 9: $<30, *, 3, 5, *>$ CSPs.

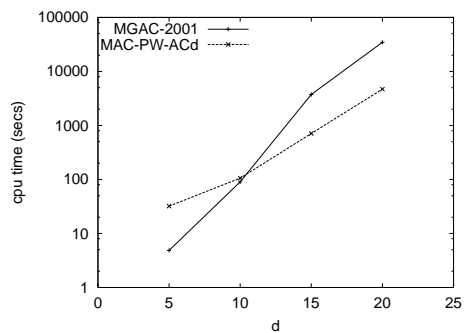

Figure 10: $<20, *, 4, 1, *>$ CSPs.

From our experiments with random problems we conjecture that the double encoding should be the preferred model for sparse problems, provided that an efficient algorithm like PW-AC is used for propagation. For CSPs with medium and high density the non-binary representation is preferable to the encodings.

**Random Problems with Added Structure** In the experiments with ternary CSPs we did not detect an advantage of the DE compared to the non-binary representation (and consequently the HVE). MAC in the DE is rarely better than MGAC-2001 (only in some cases of very tight constraints and large domain sizes), despite the use of PW-AC for propagation. Also, while MAC-PW-ACd is competitive and often faster than MGAC-2001 in sparse random problems, this result is reversed as the density increases. A basic reason for these results is that in randomly generated problems (especially ones with ternary constraints) we get many pairs of non-binary constraints that share at most one original variable. It is known (see Bacchus et al., 2002 for example) that for any such pair of constraints, the filtering achieved by AC in the DE is the same as the filtering achieved by GAC in the non-binary representation (and AC in the HVE). Therefore, AC in the DE "looses" much of its filtering power.

To validate this conjecture, we experimented with a generation model where structure is added to purely random problems. To be precise, we experimented with problems where a clique of variables is embedded in a randomly generated instance. For ternary problems any two constraints in the clique may share one or two variables. This is decided at random. For 4-ary problems any two constraints in the clique may share one, or two, or three variables. Again, this is decided at random. Table 3 compares the performance of various MAC algorithms on $<30, 10, 3, 5, *>$ and $<20, 10, 4, 1, *>$ problems of this type. For the second class we only include results for MAC-PW-ACd, which is by far the best algorithm for such problems, and MGAC-2001.

| arity | clique size | MGAC-2001 | MAC-2001 | MAC-PW-AC | MAC-2001d | MAC-PW-ACd |
|---|---|---|---|---|---|---|
| 3 | 0 | **633.50** | 45303.76 | 9205.95 | 7549.61 | 1362.31 |
| 3 | 10 | 874.07 | 21007.45 | 5349.55 | 3078.45 | **421.07** |
| 3 | 20 | 1121.39 | 1932.49 | 389.22 | 392.08 | **65.81** |
| 3 | 30 | 1598.87 | 374.03 | 48.22 | 102.30 | **5.02** |
| 4 | 0 | **90.11** | | | | 106.04 |
| 4 | 10 | 247.18 | | | | **61.12** |
| 4 | 20 | 8348.92 | | | | **322.86** |

Table 3: Average cpu times of MAC algorithms for the DE and the double encoding and MGAC for the non-binary representation of random problems with embedded cliques. All times are in seconds. Each number gives the average of 50 instances from around the phase transition region.

As we can see, the comparative results of the algorithms vary according to the size of the embedded clique. When there is no clique embedded (clique size=0) MGAC-2001 is faster than the algorithms for the binary encodings. As the clique size grows, the binary encodings, and especially the double, become more efficient. The double encoding is more effective than the DE for all clique sizes. For a large clique that covers all variables, MAC in the double encoding is many orders of magnitude faster than MGAC-2001. This huge difference is caused by the presence of many constraints that share more than one variable in the non-binary representation. In such cases filtering through the constraints between dual variables is very strong. However, much of this advantage is lost when generic algorithms are used in the encodings. Similar results occur in denser problems when this generation model is used.

### 6.2.2 Crossword Puzzles

Table 4 compares the cpu times of the two MAC algorithms in the DE and MGAC in the non-binary representation using various benchmark crossword puzzles. We do not include results for MAC in the double encoding since this particular representation of crossword puzzle generation problems is impractical. The reason for this is that for each pair of dual variables involved in a constraint, the two variables have at most one original variable in common (i.e. the letter on which the two words intersect). As explained previously, this degrades the filtering achieved by the constraints between dual variables. Such constraints in the double encoding are redundant since the same filtering can be achieved through the constraints between dual and original variables.

| puzzle | $n$ | $e$ | MGAC-2001 | MAC-2001 | MAC-PW-AC |
|---|---|---|---|---|---|
| 15.02 | 80 | 191 | **3.98** | 164.04 | 13.70 |
| 15.04* | 76 | 193 | **40.84** | 895.29 | 140.65 |
| 15.07 | 74 | 193 | **94.65** | — | — |
| 15.09 | 82 | 187 | **0.43** | — | — |
| 19.01 | 128 | 301 | **1.50** | — | — |
| 19.02 | 118 | 296 | **34.62** | — | 1028.17 |
| 19.08 | 134 | 291 | — | 33.01 | **7.76** |
| 21.02 | 130 | 295 | **2.29** | 77.51 | 11.82 |
| 21.03 | 130 | 295 | 345.96 | 273.87 | **40.57** |
| 21.08 | 150 | 365 | — | — | **9.96** |
| 21.09 | 144 | 366 | — | — | **8.87** |
| 6×6 | 12 | 36 | 9.39 | 98.49 | **6.04** |
| 7×7* | 14 | 49 | 14m | — | **12m** |
| 8×8* | 16 | 64 | **6m** | — | 9m |
| 9×9* | 18 | 81 | — | — | **128.92** |
| 10×10* | 20 | 100 | **11.81** | 474.54 | 22.78 |

Table 4: Comparison (in cpu time) between MAC algorithms for the DE and MGAC for the non-binary representation of crossword puzzles. All times are in seconds except those followed by "m" (minutes). The cpu limit was 2 hours. Problems marked by (*) are insoluble. We only include problems that were reasonably hard for either MGAC-2001 or MAC-PW-AC and at the same time were solvable within 2 hours by at least one algorithm.

From the data in Table 4 we can clearly see that MAC-PW-AC is significantly faster than MAC-2001 on all instances. The speedup offered by the use of PW-AC makes MAC in the DE competitive with MGAC in many cases where using a generic algorithm in the DE results in a clear advantage in favor of MGAC. Also, in some instances (e.g. puzzles 21.03, 21.08, 21.09), the use of PW-AC makes MAC in the DE considerably faster than MGAC. However, there are still some instances where MGAC (and consequently MHAC) finds a solution (or proves insolubility) fast, while MAC in the DE thrashes, and vice versa. Note, that only 4 of the 10 very hard 21×21 puzzles that we tried were solved by any algorithm within the time limit of two hours. MAC-PW-AC managed to solve these 4 instances relatively fast, while the other two algorithms solved only 2 of them within the cpu limit.

### 6.3 Experiments with Realistic Problems

In the next sections we present experimental results from configuration and frequency assignment problems. The aim of these experiments was to investigate the usefulness of binary encodings in realistic structured domains. We focus on the dual and double encodings which are the most promising binary encodings because of the strong propagation they can offer.

#### 6.3.1 Configuration

Configuration is an area where CSP technology has been particularly effective. A configuration problem can be viewed as trying to specify a product defined by a set of attributes, where attribute values can only be combined in predefined ways. Such problems can be modelled as CSPs, where variables correspond to attributes, the domains of the variables correspond to the possible values of the attributes, and constraints specify the predefined ways in which values can be combined. In many configuration problems the constraints are expressed extensionally through lists of allowed (or disallowed) combinations of values. Alternatively, constraints are expressed as rules which can easily be transformed into an extensional representation. Consider the following example adapted from the paper by Subbarayan, Jensen, Hadzic, Andersen, Hulgaard & Moller (2004).

**Example 6.1** The configuration of a T-shirt requires that we specify the size (small, medium, or large), the print ("Men in Black" - MIB or "Save the Whales" - STW), and the color (black, white, or red). There are the following constraints: 1) If the small size is chosen then the STW print cannot be selected. 2) If the MIB print is chosen then the black color has to be chosen as well, and if the STW print is chosen then the black color cannot be selected. This configuration problem can be modelled as a CSP with three variables $\{x_1, x_2, x_3\}$ representing size, print, and color respectively. The domains of the variables are $D(x_1) = \{small, medium, large\}$, $D(x_2) = \{MIB, STW\}$, and $D(x_3) = \{black, white, red\}$. The first constraint is a binary constraint between variables $x_1$ and $x_2$ with the following allowed tuples: $\{<small, MIB>, <medium, MIB>, <medium, STW>, <large, MIB>, <large, STW>\}$. The second constraint is a binary constraint between variables $x_2$ and $x_3$ with the following allowed tuples: $\{<MIB, black>, <STW, white>, <STW, red>\}$.

In practice, many solvers for configuration problems are able to interact with the user so that, apart from meeting the given specifications, the user's choices of values for certain attributes are also satisfied. In this study we use configuration instances to compare the non-binary representation to binary encodings on structured realistic problems. Although it would be interesting to investigate the applicability of binary encodings in an interactive configurator, such work is outside the scope of this paper.

We run experiments on five problems taken from CLib, a library of benchmark configuration problems (CLib, 2005). The first thing we noticed after encoding the five problems as binary and non-binary CSPs is that they are trivially solvable by all algorithms without backtracking. A closer look at the structure of CLibs's problems revealed the reason; their constraint graphs consist of various unconnected components. Each component consists of very few or, in some cases, a single variable. As a result, the problems are split into independent subproblems that are trivially solved by all algorithms. In order to obtain difficult instances for benchmarking, we made the graphs connected by adding some randomness.

Each of the five problems was extended by adding to it 6 variables and 8-10 constraints so that the graph became connected[8]. Table 5 shows the total number of variables and constraints in the modified problems. The added constraints were of arity 2, 3, or 4 (chosen at random) and the variables on which they were posted were selected at random, making sure that the resulting graph was connected. The looseness of each added constraint was also set at random, and finally, the allowed tuples of each constraint were chosen at random according to its looseness.

| problem | $n$ | $e$ | $arity$ | $dom$ | MGAC-2001 nodes - time | MAC-PW-AC nodes - time | MAC-PW-ACd nodes - time |
|---|---|---|---|---|---|---|---|
| machine | 30 | 22 | 4 | 9 | 535874 - 13.83 | **813 - 0.37** | 3367 - 1.64 |
| fx | 24 | 21 | 5 | 44 | 193372 - 4.10 | 92 - 0.01 | **70 - 0.01** |
| fs | 29 | 18 | 6 | 51 | 618654 - 30.43 | **41 - 0.03** | 193 - 0.05 |
| esvs | 33 | 20 | 5 | 61 | 9960160 - 332.52 | **7384 - 3.09** | 64263 - 29.86 |
| bike | 43 | 28 | 6 | 37 | 21098334 - 501.85 | **16890 - 12.23** | 112957 - 87.77 |

Table 5: Comparison of algorithms on configuration problems. *arity* and *dom* are the maximum constraint arity and maximum domain size in the problem. Run times are given in seconds.

Table 5 gives the average run times and node visits for algorithms MGAC-2001 in the non-binary representation, MAC-PW-AC in the DE, and MAC-PW-ACd in the double encoding. For each of the five benchmarks we repeatedly generated instances using the model described above. Each generated instance was solved by all three algorithms and was stored if the instance was hard for at least one algorithm. Otherwise, it was discarded. An instance was considered hard if at least one algorithm took more than one second to solve it. Table 5 reports averages over the first 50 hard instances generated from each benchmark. That is, we run 250 hard instances in total. Note that in the binary encodings all constraints of the original problem (even binary ones) were encoded as dual variables.

The experimental results of Table 5 show a very significant advantage in favor of the binary encodings compared to the non-binary representation, both in node visits and run times. The DE is clearly the most efficient model. MAC-PW-AC in the DE can be up to three orders of magnitude faster than MGAC-2001 in the non-binary representation. There was not a single instance among the 250 instances where MGAC-2001 was faster than MAC-PW-AC. The double encoding is also much more efficient than the non-binary representation. The main factor contributing to the performance of the encodings is the strong propagation that is achieved through the constraints between dual variables, which is reflected on the numbers of node visits. There is a number of reasons, related to the structure of these configuration problems, that can justify the strong performance of the encodings:

---

8. Experiments showed that these are the minimum additions that need to be made in order to get hard problems without altering the structure of the problems too much.

- The constraint graphs are very sparse. This is typical of configuration problems since, usually, an attribute of the product specification has dependencies with only a few of the other attributes.

- The constraints of high arity are very tight. Moreover, each value of the variables with large domain sizes has very few (typically one) supporting tuple in the constraints such variables participate.

- There are intersecting non-binary constraints with more than one original variable in common. As explained, and demonstrated empirically in Section 6.2, this can have a significant impact on the propagation power of AC in the dual and double encodings.

Note that the "profile" of configuration problems, as analyzed above, agrees with the conjectures we made based on results from random problems. That is, the dual and double encodings are suitable for sparse problems with tight constraints, where intersecting constraints may share more than one variable.

### 6.3.2 Frequency Assignment

Frequency assignment is an important problem in the radiocommunication industry. In such a problem there is a radio communications network in a given region consisting of a set of transmitters. Each transmitter has a position in the region, a frequency spectrum, a certain power, and a directional distribution. The aim is to assign values to some or all of the properties of the transmitters so that certain criteria are satisfied. There are various types of frequency assignment problems. In this study we consider a version of the *radio link frequency assignment problem* (RLFA). In such a problem we are given a set of links $\{L_1, \ldots, L_n\}$, each consisting of a transmitter and a receiver. Each link must be assigned a frequency from a given set $F$. At the same time the total interference at the receivers must be reduced below an acceptable level using as few frequencies as possible. These problems are typically optimization problems but for the purposes of this study we treat them as satisfaction problems.

A RLFA problem can be modelled as a CSP where each transmitter corresponds to a variable. The domain of each variable consists of the frequencies that can be assigned to the corresponding transmitter. The interferences between transmitters are modelled as binary constraints of the form $|x_i - x_j| > s$, where $x_i$ and $x_j$ are variables and $s \geq 0$ is a required frequency separation. Such a constraint restricts the frequencies that the two transmitters can simultaneously be assigned, and in that way the interference between them is minimized. This is under the realistic assumption that the closer two frequencies are the greater is the interference between them. This binary model has been used extensively to represent RLFA problems, and numerous solution methods (CSP-based or other) have been proposed. Also, RLFA has been widely used as a benchmark to test new algorithms for binary constraints (mainly AC algorithms).

It has been argued that the standard binary model of frequency assignment problems fails to capture some important aspects of real problems, such as multiple interferences, resulting in non-optimal frequency assignments (Jeavons, Dunkin, & Bater, 1998; Watkins, Hurley, & Smith, 1998; Bater, 2000; Hodge, Hurley, & Smith, 2002). As a consequence, there have been some efforts to introduce more expressive methods that utilize non-binary constraints in frequency assignment (e.g. Bater, 2000; Hodge et al., 2002). There are many types of non-binary constraints that can be considered. The following ones have received attention:

**co-channel constraints** - e.g., the frequencies assigned to $n$ transmitters are not all equal.

**adjacent-channel constraints** - e.g., the frequencies assigned to $n$ transmitters are at least one frequency apart.

**separation constraints** - e.g., the frequencies assigned to $n$ transmitters are at least $s$ frequencies apart.

Obviously, separation constraints generalize the adjacent-channel constraints. The first two types of constraints are typically very loose while the third can be tight. Separation constraints are used in densely constrained areas (representing conurbations in a region) where there is large number of links closely situated. In such cases, large separations in the frequencies of the transmitters must be imposed, resulting in tight constraints. We also consider a richer type of separation constraints: The frequencies assigned to a set of $n$ transmitters are at least $s$ frequencies apart and $n'$ transmitters among them are at least $s'(>s)$ frequencies apart from all the others. Note that some of the non-binary constraints can be equivalently decomposed into a clique of binary constraints (without introducing dual variables) resulting however in weaker propagation. An example is adjacent-channel constraints. Others cannot be equivalently expressed as a set of binary constraints unless a binary encoding is used. For example, co-channel constraints. As noted by Hodge et al. (2002), only non-binary constraints of low arity can be utilized in practice. It has been shown that in many cases such constraints are sufficient to achieve very low interferences. Constraints of higher arity may offer improvements in the quality of solutions, but they tend to slow down the solution process to an extend that solving large real problems becomes infeasible.

In the empirical study presented here we are interested in comparing models of RLFA-type problems with non-binary constraints to the corresponding binary encodings and not in devising new efficient methods for solving RLFA problems. Since the available RLFA benchmarks follow the standard binary approach, to test the algorithms we generated non-binary problems by placing variables, corresponding to links, on a grid following the typical RLFA structure. That is, the problems consist of several groups of closely situated variables plus a few constraints that connect these groups. For example, such structures are depicted in Figure 11. This corresponds to the constraint graph of a binary RLFA problem which typically consists of a set of cliques (or near-cliques) of binary constraints and a small number of constraints connecting the various cliques (e.g. the benchmarks of Cabon, De Givry, Lobjois, Schiex, & Warners, 1999). From the binary encodings we only considered the double since dual variables can have very large domains, which makes the DE inefficient.

Indicative results of the experiments we run are depicted in Table 6. In these experiments we posted only low-arity (i.e. 3-ary to 5-ary) separation constraints, as shown in Figure 11, and compared the performance of algorithm MGAC-2001 on the non-binary model of the problems to the performance of MAC-PW-ACd on the double encoding of the problems. We tried two implementations of MGAC-2001; one that utilizes specialized propagators for the separation constraints (written as functions), and another that operates on the extensional representation of the constraints. The first implementation was generally faster, so all the results of MGAC-2001 presented below refer to the intentional implementation. The double encoding was built by translating the separation constraints into lists of allowed tuples in a preprocessing step.

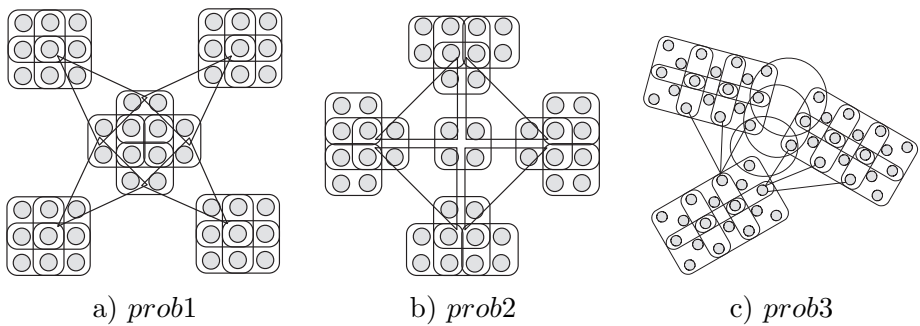

a) *prob*1  b) *prob*2  c) *prob*3

Figure 11: Examples of RLFA problems with non-binary separation constraints.

| problem | $n$ | $e$ | $arity$ | MGAC-2001 nodes - time | MAC-PW-ACd nodes - time |
|---|---|---|---|---|---|
| *prob*1 (easiest) | 48 | 25 | 4 | 583 - 1.51 | **48 - 0.36** |
| *prob*1 (median) | 48 | 25 | 4 | 18161268 - 39518.71 | **50 - 0.34** |
| *prob*1 (hardest) | 48 | 25 | 4 | — | — |
| *prob*2 (easiest) | 44 | 21 | 4 | **56474 - 26.46** | 172071 - 219.06 |
| *prob*2 (median) | 44 | 21 | 4 | **304676 - 169.01** | 305588 - 504.06 |
| *prob*2 (hardest) | 44 | 21 | 4 | **53654514 - 4220.21** | 68771412 - 24133.45 |
| *prob*3 (easiest) | 54 | 24 | 5 | 103 - 13.45 | **0 - 6.42** |
| *prob*3 (median) | 54 | 24 | 5 | **2134 - 14.14** | 2569 - 53.18 |
| *prob*3 (hardest) | 54 | 24 | 5 | 4194 - 115.12 | **0 - 5.64** |
| *prob*4 (easiest) | 68 | 38 | 5 | **70 - 1.21** | 72 - 7.24 |
| *prob*4 (median) | 68 | 38 | 5 | — | **90 - 8.34** |
| *prob*4 (hardest) | 68 | 38 | 5 | — | — |
| *prob*5 (easiest) | 100 | 58 | 5 | 294 - 5.29 | **0 - 2.42** |
| *prob*5 (median) | 100 | 58 | 5 | 96106 - 522.64 | **99 - 2.81** |
| *prob*5 (hardest) | 100 | 58 | 5 | — | **104 - 13.45** |

Table 6: Comparison of algorithms on RLFA problems with separation constraints. *arity* is the maximum constraint arity. Run times are given in seconds. A dash (—) is placed wherever the algorithm did not finish its run within 12 hours of cpu time.

Table 6 reports results from a total of 50 instances created using five different constraint graph topologies. All variables have domains of 20 or 25 values. The number of allowed tuples for the constraints varied from around 50 for very tight constraints to several thousands for looser ones, according to the frequency separation imposed by parameters $s$ and $s'$ of the separation constraints. These parameters were set at random for each constraint, making sure that very loose constraints were not generated. For example, for a 4-ary basic separation constraint on variables with domain size 20, $s$ was at least 3 (giving 7920 allowed tuples) and at most 5 (giving 120 allowed tuples).

*prob*1, *prob*2, and *prob*3 refer to problems having the topologies shown in Figure 11. *prob*4 consists of three groups of variables, similar to the ones of *prob*3, arranged in a chain-like structure. Finally, instances of *prob*5 consist of randomly generated groups of variables; each one having 8-10 variables and 3-5 3-ary to 5-ary constraints. These groups are interconnected according to their topological distance (i.e. constraints are posted on variables of nearby groups). All instances of *prob*1-*prob*4 have fixed topology. For each topology a set of instances was created by changing the type of the constraints. For example, two instances having the topology of *prob*1 may differ in the type of separation constraints (basic or richer) they include. Also, the frequency separations $s$ and $s'$ imposed by a constraint may differ. Instances of *prob*5 may also differ in their constraint graph topology. We report node visits and run times of the easiest, median, and hardest instance for each topology, with respect to the performance of MGAC[9]. The hardest instances were the same for both the encoding and the non-binary representation (except for *prob*3), while the easiest and median instances were sometimes different.

In Table 6 we can see that there can be very substantial differences in favor of the double encoding. Many instances were solvable in the double encoding with no or very little backtracking while MGAC-2001 thrashed. This is mainly due to the large number of interleaved constraints sharing more than one variable, which boosts propagation in the double encoding. The performance of the algorithms seems to be heavily dependent on the topology of the problems. For example, on instances of *prob*2 the non-binary representation was much more efficient than the double encoding. It seems that in this particular class of problems heuristic choices were misled by the propagation achieved in the double encoding. We have not been able to come up with a satisfactory explanation as to why this occurred in this particular topology.

Finally, to investigate the effect that the presence of loose constraints of higher arity has, we run experiments where 8-ary adjacent-channel constraints were posted between variables further apart in the graph, in addition to the separation constraints. In this case using the double encoding to model all the constraints in the problems was infeasible due to the spatial requirements. For example, trying to generate the allowed tuples of a single 8-ary adjacent-channel constraint consumed all the memory of the system. Therefore, we compared algorithm MGAC-2001 on the non-binary model to a MAC algorithm that runs on a hybrid model where the tight separation constraints are modelled using the double encoding and the loose adjacent-channel constraints are kept in the intentional non-binary representation. Table 7 reports results from a total of 30 instances created using the graph topologies of $prob1$, $prob2$ and $prob5$ with the addition of four 8-ary adjacent-channel constraints in each instance. The hybrid model is more efficient on instances of $prob1$ and $prob5$ because of the strong propagation achieved through the binary encoding of the tight constraints. The non-binary model is better on instances of $prob2$ where it seems that propagation through the binary encoding results in bad heuristic choices.

---

9. To create the instances we varied the type of the constraints and the values of parameters $s$ and $s'$ until non-trivial problems were generated. We consider as trivial problems that are arc inconsistent or solvable with no backtracking.

| problem | $n$ | $e$ | $arity$ | MGAC-2001 nodes - time | MAC-hybrid nodes - time |
|---|---|---|---|---|---|
| $prob1$ (easiest) | 48 | 25 | 8 | **106 - 20.88** | 50 - 47.60 |
| $prob1$ (median) | 48 | 25 | 8 | 5078 - 1201.98 | **84 - 195.43** |
| $prob1$ (hardest) | 48 | 25 | 8 | — | — |
| $prob2$ (easiest) | 44 | 21 | 8 | **647 - 192.64** | 1019 - 308.84 |
| $prob2$ (median) | 44 | 21 | 8 | **80245 - 17690.12** | — |
| $prob2$ (hardest) | 44 | 21 | 8 | — | — |
| $prob5$ (easiest) | 100 | 58 | 8 | 76 - 45.92 | **0 - 22.19** |
| $prob5$ (median) | 100 | 58 | 8 | 22785 - 3230.47 | **99 - 78.41** |
| $prob5$ (hardest) | 100 | 58 | 8 | — | **18447 - 4233.50** |

Table 7: Comparison of algorithms on RLFA problems with separation and adjacent-channel constraints. MAC-hybrid corresponds to a MAC algorithm that runs on the hybrid model.

### 6.4 Discussion

In this section we summarize the results of our experimental studies and draw some conclusions regarding the applicability of the encodings, based on our theoretical and experimental analysis.

**Hidden Variable Encoding** As theoretical results suggested, and empirical results confirmed, solving problems in the HVE using algorithms that only instantiate original variables is essentially analogous to solving the non-binary representation directly. All the commonly used algorithms for non-binary problems can be applied, with adjustments, to the HVE, and vice versa. When such algorithms are used, the HVE offers some (moderate) computational savings compared to the non-binary representation, especially in sparse problems. These savings are due to the ability of the AC algorithm in the HVE to detect inconsistencies earlier than the corresponding GAC algorithm in the non-binary representation. Therefore, we conjecture that the HVE is applicable in sparse non-binary problems where the constraints are extensionally specified. In other cases, the HVE is either less efficient in run times than the non-binary representation (e.g. dense problems), or building the HVE adds space overheads that are not justified by the marginal gains in search effort. Additionally, there is not enough empirical evidence to suggest that the essential difference between search algorithms for the HVE and the non-binary representation, i.e. the ability of the former to branch on dual variables, can make the HVE significantly more efficient in some class of problems. This, coupled with the fact that any benefits gained by instantiating dual variables can be maximized if the double encoding is used instead of the HVE, limits the applicability of such algorithms.

**Dual and Double Encodings** The DE and the double encoding have the advantage of strong filtering through the constraints between dual variables. We showed that this advantage can be exploited by a low cost specialized algorithm, such as PW-AC, to make the DE competitive and often significantly better than the non-binary representation in several sparse CSPs, such as crossword puzzle generation and configuration problems. For dense CSPs, the DE does not pay off because either the spatial requirements make its use infeasible, or if this is not the case, the advantages offered are outweighed by the overhead of updating the domains of dual variables. The same holds for CSPs containing constraints of large arity unless they are very tight (as in crossword puzzles).

Algorithms for the double encoding demonstrate especially promising performance. When many non-binary constraints that share more than one variable are present in a problem then MAC in the double encoding can exploit the benefits of both the variable ordering heuristic, borrowed from the non-binary representation, and the stronger filtering, borrowed from the DE, to outperform the other representations. This was demonstrated in problems with such structure (random and also frequency assignment - like). This is also the case in the "still-life" problem, which explains the success of the double encoding[10]. In addition, the double encoding offers the interesting potential of hybrid models where certain constraints are encoded into binary and others are kept in the non-binary representation based on certain properties of the constraints. To be precise we can benefit by encoding constraints that are either naturally specified in extension, or have relatively low arity and are tight. This was demonstrated in various domains. Most notably, in the frequency assignment problems where the double encoding (or a hybrid one) payed off in most cases, although the constraints in such problems are naturally defined intentionally.

## 7. Related Work

Although, the DE was proposed in 1989 (Dechter & Pearl, 1989) and the HVE in 1990 (Rossi et al., 1990), the first substantial effort towards evaluating their efficiency was carried out in 1998 (Bacchus & van Beek, 1998). In that work, Bacchus and van Beek compared theoretically and empirically the FC algorithm in the two encodings against FC for non-binary CSPs. Also, they introduced FC+, a specialized algorithm for the HVE. The algorithms compared by Bacchus and van Beek were the simplest versions of FC; hFC0 and hFC1 (i.e. FC+) for the HVE, and nFC0 for the non-binary representation. We extend that work by studying various recent and more advanced versions of FC.

Following Bacchus and van Beek (1998), Stergiou and Walsh made a theoretical and empirical study of AC in the encodings (Stergiou & Walsh, 1999). It was proved that AC in the HVE is equivalent to GAC in the non-binary representation, while AC in the DE is stronger. In the small experimental study included in the paper by Stergiou & Walsh (1999), MAC for the HVE, DE, and double encoding was compared to MGAC in the non-binary representation of some crossword puzzles and Golomb rulers problems. Results showed an advantage for the non-binary representation and the HVE, but it is important to note that all the MAC algorithms used generic inefficient algorithms to enforce AC.

---

10. Although the title of Smith's paper (2002) refers to the DE, the model of the "still-life" problem used is based on the double encoding.

Smith, Stergiou & Walsh (2000) performed a more extensive experimental comparison of MAC in the HVE and the double encoding, and MGAC in a non-binary model of the Golomb rulers problem. However, again the MAC algorithms in the encodings used a generic algorithm to enforce AC. As a result they were outperformed by MGAC in the non-binary model.

Beacham et al. (2001) compared the performances of different models, heuristics, and algorithms for CSPs using crossword puzzle generation problems as benchmarks. Among the models that were compared was the HVE, the DE and the non-binary representation. Once again, the algorithms that were applied in the encodings were generic algorithms. For example, two of the implemented algorithms for the DE were MAC that uses AC-3 for propagation and MAC that uses AC-7. Both these algorithms suffer from the very high complexity of AC propagation. As demonstrated, the use of algorithm PW-AC for propagation can significantly enhance the performance of MAC in crossword puzzle problems.

Bacchus et al. (2002) presented an extensive theoretical study of the DE and the HVE. Among other results, polynomial bounds were placed on the relative performance of FC and MAC in the two encodings and the non-binary representation, or it was shown that no polynomial bound exists. For example, it was shown that FC in the HVE (i.e. hFC0 in the terminology we use) is never more than a polynomial factor worse than FC in the DE, but FC in the DE can be exponentially worse than FC in the HVE. Also, FC in the non-binary representation (i.e. nFC0 in the terminology we use) can be exponentially worse than FC in the HVE, and vice versa. We add to these results by analyzing the performance of various more advanced algorithms for the HVE and the double encoding.

Smith modelled the problem of finding a maximum density stable pattern "still-life" in Conway's game of Life using MAC in the double encoding with remarkable success, compared to other constraint programming and integer programming approaches (Smith, 2002). The MAC algorithm was implemented using the *Table* constraint of ILOG Solver. This constraint implements the generic AC algorithm of Bessiére & Régin (1996a), which is very expensive when used in the DE because of its high time complexity. We believe that the results presented by Smith can be further improved if MAC-PW-AC is used instead.

## 8. Conclusion

In this paper we studied three binary translations of non-binary CSPs; the hidden variable encoding, the dual encoding, and the double encoding. We showed that the common perception that standard algorithms for binary CSPs can be used in the encodings of non-binary CSPs suffers from flaws. Namely, standard algorithms do not exploit the structure of the encodings, and end up being inefficient. To address this problem, we proposed specialized arc consistency and search algorithms for the encodings, and we evaluated them theoretically and empirically. We showed how arc consistency can be enforced on the hidden variable encoding of a non-binary CSP with the same worst-case time complexity as generalized arc consistency on the non-binary representation. We showed that the structure of constraints in the dual encoding can be exploited to achieve a much lower time complexity than a generic algorithm. Empirical results demonstrated that the use of a specialized algorithm makes the dual encoding significantly more efficient. We showed that generalized search algorithms for non-binary CSPs can be relatively easily adjusted to operate in the hidden variable encoding. We also showed how various algorithms for the double encoding can be designed. These algorithms can exploit the properties of the double encoding (strong filtering and branching on original variables) to achieve very good results in certain problems. Empirical results in random and structured problems showed that, for certain classes of non-binary constraints, using binary encodings is a competitive option, and in many cases, a better one than solving the non-binary representation.

## Acknowledgements

We would like to thank Panagiotis Karagiannis, Nikos Mamoulis, and Toby Walsh for their help in various stages of this work. We would also like to thank the anonymous reviewers of an earlier version of this paper for their very useful comments and suggestions.

## Appendix A

As explained, the main difference between the AC algorithm for the HVE and the corresponding GAC algorithm is the fact that the AC algorithm has to update the domains of the dual variables as well as the original ones. This incurs a time overhead, but as we will show, deleting values from dual variables can help propagation discover domain wipe-outs in arc inconsistent problems faster.

**Proposition 8.1** Let $P$ be a non-binary CSP. Assume that after generalized arc consistency is applied in $P$, there is no domain wipeout in the resulting problem. Enforcing arc consistency in the hidden variable encoding of $P$ using HAC requires the same number of consistency checks as enforcing generalized arc consistency in $P$ using GAC-2001, assuming the two algorithms follow the same ordering of variables and values when looking for supports and propagating deletions.

**Proof:** First, consider that if no domain wipeout in any variable (original or dual) occurs then the two algorithms will add constraints (dual variables) to the stack and remove them for revision in exactly the same order. Therefore, we only need to show that if a value is deleted from a variable during the revision of a constraint or finds a new support in the constraint then these operations will require the same number of checks in both representations. Assume that in the non-binary version of the algorithm value $a$ is deleted from the domain of variable $x_i$ because it has no support in constraint $c$. If $|T|$ is the number of allowed tuples in $c$ then determining the lack of support will require $|T| - currentSupport_{x_i,a,c}$ checks, one for each of the tuples in $c$ that have not been checked yet. If the value is not deleted but finds a new support $\tau$, with $\tau > currentSupport_{x_i,a,c}$, then $\tau - currentSupport_{x_i,a,c}$ checks will be performed. In the HVE, $x_i$ will be processed in the same order as in the non-binary version and we will require $|T| - currentSupport_{x_i,a,v_c}$ or $\tau - currentSupport_{x_i,a,v_c}$ checks depending on the case. Obviously, $currentSupport_{x_i,a,c}$ is the same as $currentSupport_{x_i,a,v_c}$ since a tuple in $c$ corresponds to a value in $v_c$, and therefore, the same number of checks will be performed in both representations. □

Samaras & Stergiou

**Proposition 8.2** Let $P$ be a non-binary CSP. Assume that the application of generalized arc consistency in $P$ results in a domain wipeout. Algorithm HAC applied in the hidden variable encoding of $P$ discovers the domain wipeout with at most the same number of consistency checks as algorithm GAC-2001 in the non-binary representation, assuming the two algorithms follow the same ordering of variables and values when looking for supports and propagating deletions.

**Proof:** In any CSP, arc inconsistency in detected when the domain of a variable is wiped out while applying AC. In the HVE of a non-binary CSP, arc inconsistency is detected when the domain of an original variable is wiped out or (crucially) when the domain of a dual variable is wiped out. The second possibility can make an AC algorithm that operates in the HVE more efficient than the corresponding GAC algorithm. To prove this consider an arc inconsistent non-binary problem. Assume that the domain of original variable $x_i$ is wiped out during the processing of constraint $c$ which is encoded as a dual variable $v_c$ in the HVE. Until the point where function *Revise* is called with $x_i$ and $c$ as arguments, there is no inconsistency and according to Proposition 8.1 the GAC algorithm and the AC algorithm on the HVE will perform the same number of consistency checks. Assume that there are $j$ values left in $D(x_i)$ before the call to *Revise*. In function *Revise* we will unsuccessfully look for a support for each of the $j$ values. If $|T|$ is the number of allowed tuples in $c$ then, for each value $a \in D(x_i)$, this will require $|T| - currentSupport_{x_i,a,c}$ checks for the GAC algorithm and $|T| - currentSupport_{x_i,a,v_c}$ checks for the AC algorithm. Since $|T| - currentSupport_{x_i,a,c} = |T| - currentSupport_{x_i,a,v_c}$, the two algorithms will perform the same number of consistency checks to detect the domain wipeout.

The following example demonstrates that HAC may discover the inconsistency with less checks. Consider a problem with variables $x_1$, $x_2$, $x_3$, $x_4$ which have domains $\{0,1\}$, $\{0,1\}$, $\{0,\ldots,9\}$, and $\{0,1\}$, respectively. There are two constraints, $c_1$ and $c_2$, with $vars(c_1) = \{x_1, x_2, x_3\}$ and $vars(c_2) = \{x_1, x_2, x_4\}$ respectively. Value 0 of $x_2$ is supported in $c_1$ by tuples that include the assignment $(x_1, 1)$. Value 0 of $x_1$ is supported in $c_2$ by tuples that include the assignment $(x_2, 0)$. Constraint $c_2$ allows only tuples that include the assignment $(x_2, 0)$. Values $0, \ldots, 9$ of $x_3$ are supported in $c_1$ by tuples that include $(x_2, 0)$ and by tuples that include $(x_2, 1)$. Now assume that variable $x_1$ is instantiated to 0, which means that the deletion of 1 from $D(x_1)$ must be propagated. In the HVE, we will first delete all tuples that include the value $(x_1, 1)$ from dual variables $v_{c_1}$ and $v_{c_2}$. Then, we will add dual variables $v_{c_1}$ and $v_{c_2}$ to the stack, remove them, and revise all original variables connected to them. Assuming that $v_{c_1}$ is removed first, value 0 of $x_2$ will have no support in $v_{c_1}$ so it will be deleted. As a result, we will delete all tuples from dual variable $v_{c_2}$ that include the pair $(x_2, 0)$. This means that the domain of $v_{c_2}$ will be wiped out. In the non-binary representation, we will proceed in a similar way and perform the same number of checks until 0 is deleted from $x_2$. After that deletion the algorithm will look for supports in $c_1$ for value 1 of $x_2$ and all values of $x_3$. This will involve checks that are avoided in the HVE. The inconsistency will be discovered later when we process constraint $c_2$ and find out that value 1 of $x_2$ has no support in $c_2$ resulting in the domain wipeout of $x_2$. □